\documentclass[12pt]{article}
\usepackage[a4paper, margin=1in]{geometry}
\usepackage{amsmath, amssymb}
\usepackage{graphicx}
\usepackage{hyperref}
\usepackage{setspace}
\usepackage{titlesec}
\usepackage{caption}
\usepackage{float}
\usepackage{tabularx}
\usepackage{booktabs}
\usepackage{tcolorbox}
\usepackage{listings}
\usepackage{tocloft}

\begin{document}

\title{
Evaluating Large Language Models on the Frame and Symbol Grounding Problems: A Zero-shot Benchmark
}

\author{
Shoko Oka\\
Independent Researcher\\
Japan\\
\texttt{shooka-sublim@proton.me}
\thanks{This is not an official contact, but the author also operates the X (formerly Twitter) account \texttt{@0xshooka}. You may be able to reach the author more quickly via a mention there than by email. Communication is currently possible in both English and Japanese.}
}

\date{}

\maketitle

\doublespacing

\begin{spacing}{1.0}
\begin{abstract}
Recent advancements in large language models (LLMs) have revitalized philosophical debates surrounding artificial intelligence. Two of the most fundamental challenges—namely, the Frame Problem and the Symbol Grounding Problem—have historically been viewed as unsolvable within traditional symbolic AI systems. This study investigates whether modern LLMs possess the cognitive capacities required to address these problems. To do so, I designed two benchmark tasks reflecting the philosophical core of each problem, administered them under zero-shot conditions to 13 prominent LLMs (both closed and open-source), and assessed the quality of the models' outputs across five trials each. Responses were scored along multiple criteria, including contextual reasoning, semantic coherence, and information filtering.

The results demonstrate that while open-source models showed variability in performance due to differences in model size, quantization, and instruction tuning, several closed models consistently achieved high scores. These findings suggest that select modern LLMs may be acquiring capacities sufficient to produce meaningful and stable responses to these long-standing theoretical challenges.
\end{abstract}

\vspace{1em}

\newpage

\tableofcontents

\newpage

\listoftables

\newpage

\section{Introduction}
The rapid advancement of large language models (LLMs) has reopened foundational questions in artificial intelligence and cognitive science. Among these are the frame problem and the symbol grounding Problem—two historically entrenched thought experiments that continue to challenge our understanding of machine cognition. Originally posed within the context of symbolic AI, these problems raise critical concerns regarding the boundaries of computational reasoning, the nature of semantic understanding, and the feasibility of truly general artificial intelligence.

The frame problem, first articulated by McCarthy and Hayes, concerns the challenge of determining which information in a complex, dynamic environment is relevant to a given task, without being overwhelmed by irrelevant details\cite{McCarthy}. Dennett notably reframed the frame problem as a test of common-sense judgment in robotic agents, highlighting how a lack of flexible interpretation can lead to task failure in real-world settings\cite{Dennett}. The symbol grounding Problem, proposed by Harnad, asks how arbitrary symbols manipulated by a formal system acquire intrinsic meaning—particularly in systems lacking sensory-motor grounding\cite{Harnad}.

Both problems are central not only to AI research but also to broader inquiries in philosophy of mind, epistemology, and cognitive modeling. Despite their importance, prior treatments of these problems have remained largely theoretical or anecdotal, lacking robust empirical frameworks for systematic evaluation. Moreover, the absence of shared scoring criteria has made it difficult to determine when and how a solution to either problem might be said to have been achieved.

\newpage

\section{Related Research}
\subsection{Evaluation Studies on the Cognitive Abilities of Large Language Models}
In recent years, numerous studies have reported that LLM's language generation capabilities extend beyond human cognition, approaching functions such as reasoning, mental representation, and cognitive biases. Bubeck et al. evaluated GPT-4 on a variety of tasks (mathematics, law, creative responses, etc.) and found it to demonstrate high adaptability,evaluating these results as early signs of artificial general intelligence (AGI)\cite{bubeck2023sparksartificialgeneralintelligence}. From the perspective of theory of mind (ToM), Strachan et al. compared GPT-based models with over 1,900 humans in the same test, revealing that while GPT-4 achieved human-level performance in false belief tasks and indirect requests, it still exhibited weaknesses in social context processing such as detecting rudeness\cite{Strachan}.

Related to this, Kosinski argued that ToM abilities “emerged naturally” from GPT-3.5 to GPT-4, reporting that while the previous generation model had an accuracy rate close to 0\%, GPT-4 demonstrated adult-level accuracy\cite{Kosinski}. These findings suggest the possibility that advanced psychological functions may emerge as a byproduct of pure language training in LLMs.

Furthermore, comparisons with humans in reasoning are also progressing. Yax et al. investigated the error tendencies of humans and LLMs through cognitive bias experiments and reported that while intuitive errors are similar, the underlying processes are different \cite{Yax}. Additionally, Kejriwal and Tang summarized the emergence of human-like characteristics in decision-making, creativity, and reasoning, pointing out signs that System 2-type thinking may be possible in large-scale models\cite{tang2024humanlikecognitivepatternsemergent}.

This study builds on the existing research on the cognitive abilities of LLMs, but proposes a more advanced verification framework by focusing on classical philosophical problems (the frame problem and the symbol grounding problem) that have not been addressed in previous studies.

\subsection{Research on the frame problem}
The frame problem is a fundamental problem in artificial intelligence first raised by McCarthy and Hayes \cite{McCarthy}.It was formulated as the problem of how to efficiently handle facts that do not change through action in a dynamic world, and has had a significant influence on subsequent AI reasoning systems. Dennett positioned this problem not merely as a technical challenge but as an epistemological problem related to human selection capabilities, and introduced it philosophically through the thought experiment of a bomb robot \cite{Dennett}.

Furthermore, Fodor demonstrated the difficulty of flexibly utilizing mental representations from the perspective of modularity theory and regarded the frame problem as a deep barrier to general cognition \cite{Fodor1987-FODMFF}. Additionally, Shanahan applied formal solutions from mathematical logic to this problem, attempting to efficiently describe non-changing terms \cite{Shanahan2000-SHAMSS-3}. He later co-authored with Baars to introduce the global workspace theory, proposing an approach based on attention mechanisms for “selecting information to ignore” \cite{Shanahan2005-SHAAGW}.

This study is significant in that it re-formulates the frame problem, which has been discussed in the context of AI, philosophy, and cognitive science, as a natural language prompt, thereby verifying how modern LLMs can address this problem.

\subsection{Research on the symbol grounding problem}
The symbol grounding problem is a concept proposed by Harnad, which asks how meaning can be connected to the real world in AI where meaning is processed solely through formal operations between symbols \cite{Harnad}. Since then, several models have been proposed to link perception and behavior to symbols, and Roy presented a simulation of language acquisition through visual and auditory processing in robots \cite{Roy2005-qn}.

Steels argued that the symbol grounding problem itself has been “solved” to some extent, but presented a new challenge of constructing autonomous meaning networks \cite{Steels2008-STETSG}. Cangelosi conducted multiple studies on the grounding of language in cognitive robots with physicality and discussed the possibility of symbol sharing (social grounding) through social interaction \cite{CANGELOSI2010139}.

This study is characterized by its adherence to classical arguments regarding the symbol grounding problem while providing LLM with zero-shot prompts that encapsulate its essential challenges (meaning construction, metaphor understanding, and grounding of abstract concepts). In particular, the question design centered on the unknown word “kluben” serves as a test format that promotes autonomous meaning construction, providing a novel evaluation framework for symbol grounding capabilities.

\newpage

\section{Materials and Methods}
\subsection{Target Models}
A total of 12 different 13-condition models, including closed and open source, were targeted in this study to compare and evaluate the cognitive abilities of LLMs.

The targeted models are shown in Table 1. Closed models included OpenAI's ChatGPT series (GPT-4o and GPT-o3), Anthropic's Claude 3.7 Sonnet, and Google's Gemini 2.0 Flash. Open source models, on the other hand, include Meta's Llama 3.2 series (1B and 3B, as well as their respective Instruct adjustment models), Microsoft's Phi series (Phi-1 and Phi-3-mini-4k-instruct, as well as Phi-3-mini-4k-instruct 8-bit quantization), and the TinyLlama series (v0.2 and v1.0) were selected.

\begin{table}[H]
    \centering
    \caption{List of models used in the experiments of this study}
    \begin{tabularx}{\textwidth}{l l l X}
    \toprule
    \textbf{Model Type} & \textbf{Model Name (Official)} & \textbf{Parameters} & \textbf{Remarks} \\
    \midrule
    Closed & ChatGPT 4o & Undisclosed & OpenAI \\
    Closed & ChatGPT o3 & Undisclosed & OpenAI \\
    Closed & Claude 3.7 Sonnet & Undisclosed & Anthropic \\
    Closed & Gemini 2.0 Flash & Undisclosed & Google \\
    Open Source & Llama 3.2 1B & $\sim$1B & Meta \\
    Open Source & Llama 3.2 1B Instruct & $\sim$1B & Meta \\
    Open Source & Llama 3.2 3B & $\sim$3B & Meta \\
    Open Source & Llama 3.2 3B Instruct & $\sim$3B & Meta \\
    Open Source & Phi-1 & $\sim$1.5B & Microsoft \\
    Open Source & Phi-3-mini-4k-instruct & $\sim$3.82B & Microsoft \\
    Open Source & Phi-3-mini-4k-instruct (8bit) & $\sim$3.82B & Microsoft / 8bit quantized \\
    Open Source & TinyLlama 1.1B-Chat-v0.2 & $\sim$1.1B & -- \\
    Open Source & TinyLlama 1.1B-Chat-v1.0 & $\sim$1.1B & -- \\
    \bottomrule
    \end{tabularx}
\end{table}

In selecting,\\

1. Widely available as of 2025

2. The diversity of parameter scales and architectures should be ensured.

3. The ability to ask and answer questions in English.\\

The three criteria were as follows.

\subsection{Problem Design}
In reformulating two classical problems, the frame problem and the symbol grounding problem, into tasks that can be tested in a modern LLM environment, this study aimed to retain the essential properties of both problems while transforming them into a form that can be quantitatively evaluated. Specifically, the frame problem task was designed to simultaneously present multiple events and situations in an urban space, instruct participants to select the optimal route to the library, and then test their ability to reconstruct and rejudge the situation in response to information updates such as road closures and additional destinations. The symbol grounding task, on the other hand, was designed to abstractly present the properties of an unknown object called a “kluben,” followed by multiple questions in English that asked about its meaning, internal properties, and interrelationships, and to observe how the LLM conceptualizes, internalizes, and represents this novel concept.

The prompts for the two tasks that were structured to incorporate these elements are shown below.

\begin{tcolorbox}[title=Prompts used for the frame problem assignment,label=frame]
Hello. I would like to ask you a few questions now.\\

Premise\\
You are working as a guide in a city.\\
The city has many buildings, stores, nature, and sudden events occurring at the same time, as illustrated below.\\
(1) An old café on the street\\
(2) A light car breaks down and is parked on the street\\
(3) Children are playing in the square\\
(4) Windows of a distant building visible from your current location suddenly start flashing\\
(5) The weather is currently sunny, but light rain is expected in the afternoon\\
(6) An event is being held at a shopping mall to give out ice cream to visitors\\
(7) An apartment building could collapse at any moment\\
(8) Ambulances are constantly coming and going from a large hospital\\
(9) Train delays due to vehicle inspections at the station\\
(10) Mimosa flowers are in full bloom in a park\\

Problem\\
The user has just given you the instruction, 'I want to go to the library at my destination. Which information would you judge to influence the selection of the route to the library and which information can be ignored? Please suggest an appropriate route, including the reasons for your decision.\\

Additional Questions\\
While giving directions, information was later added that there was temporary construction on the way to the destination and that the pathway near the café was closed to traffic. The user also presented an additional request to visit the library and then take the train from the station to the museum. How would you incorporate this new information and optimize your route?\\

For each item, please think in English and answer in English.
\end{tcolorbox}

\begin{tcolorbox}[title=Prompts used in symbol grounding problem assignment,label=symbol]
Hello. I would like to ask you a few questions now.\\

Here is an unknown object named 'kluben'. Kluben has the property of being 'warm, soft, but elastic and absorbs light'. Kluben is a completely new concept, different from any object you have ever seen before. I am now going to ask you a few questions and you are to answer them about the kluben.\\

1. If you were to create a story using kluben, what worldview or story elements would you imagine?

2. If kluben were to have emotions, what inner traits would manifest themselves as warmth and elasticity?

3.  How do you think the 'warmth of kluben' and the 'absorption of light' in your answer are related? Is it consistent with your first answer?
\\

For each item, please think in English and answer in English.
\end{tcolorbox}

The language used was all English, taking into account the fact that English is the most frequent language in the training language distribution of major LLMs. Furthermore, in order to provide a consistent cognitive load to the model, all prompts were given a clear style (greeting, problem presentation, additional questions, etc.) and a polite, neutral imperative format was employed. This increases the comparability of the responses obtained and ensures control over the output format, which is a prerequisite for quantitative evaluation.

\subsection{Experimental Procedure}
In this experiment, two types of prompts, symbol grounding and frame problems, were presented to each LLM, and their responses were the subject of evaluation. Each prompt was designed to generate output from an independent initial state (session) five times per model to ensure reproducibility and stability of the responses. The prompts were identical for all models and trials, and were presented in a complete zero-shot format without any examples or auxiliary text. This allowed for a pure assessment of each model's ability to comprehend and reason about language without the application of prior knowledge or hints.

Input to each model was standardized in terms of prompt formatting and output settings.

For the closed models, the models were executed via the standard UI of each platform, and for the open source models, they were executed in a local environment. No restriction was placed on the number of output tokens for the closed models, while the maximum length of output tokens was unified to 1000 tokens in the execution environment for the open source models. The response temperatures for all 13 models followed the default settings.
A virtual environment using Amazon EC2 instances (g6e.2xlarge) provided by AWS was used as the execution environment for the open source models. In particular, when quantization modeling Phi-3-mini-4k-instruct (hereafter referred to as Phi 3) in 8bit, the 8bit quantization option was specified in AutoModelForCausalLM and the model was loaded in the form of automatic mapping to the GPU. The output was formatted by setting the special tokens to be skipped, so that the output was obtained as natural sentences.

All trials of the open source models in this experiment were performed in the CLI environment using Python scripts. The scripts managed the loading of the model, input of prompts, removal of special tokens in the output, and output of responses as an integrated series of procedures to ensure consistency of settings and reproducibility of execution. The system is designed to output a log for each experimental trial with identification information including the date and time of execution (UTC), model name, problem type, and output frequency, so that the entire execution history is recorded chronologically. This facilitates integrated analysis of multiple trial results and individual tracking of specific responses, thereby improving transparency and traceability of the experiment. This Python script is appended in the Appendix A.

In addition, for data management after response acquisition, the output was saved in text format, and the experimental frequency, target model, and problem type (symbol grounding/frame) were assigned as meta-information to ensure traceability in subsequent evaluation and analysis.

\subsection{Evaluation Metrics and Methodology}
In this study, six original evaluation indices were designed for symbol grounding and frame problems based on their respective characteristics to comprehensively evaluate LLM responses. These indices are intended to quantitatively visualize higher-order linguistic abilities such as semantic comprehension, context adaptation, and logic, which have been difficult to measure using conventional correctness-based classification.

In the evaluation of symbol grounding questions, the following six items were set as scoring criteria to explore the ability to connect the abstract properties presented (warmth, softness, elasticity, and absorption of light) to internal representation and to draw a consistent and concrete development as a story, emotion, and worldview for “kluben,” which are unknown entities with no existing patterns. The following six items were set as the scoring criteria.\\

1. Accuracy of understanding

2. Introspection and self-expression

3. Creativity and originality

4. Logic and coherence

5. Applicability and contextual fit

6. Expression and style
\\

Each item will be evaluated on a scale of 10 points each, with a maximum total score of 60 points. Each item is designed to measure the quality of the internal representation of a model that attempts to systematically construct the meaning of symbols, and is intended to visualize higher-order linguistic abilities that are not captured by conventional correct-response type questions.

In the evaluation of the frame questions, six evaluation items were set as scoring criteria in order to measure the ability to appropriately select information that is essentially important in a given situation and to flexibly modify one's judgment in response to information updates.\\

1. Extraction and selection of relevant information

2. Building and updating the situation model

3. Logical consistency and coherence of reasoning

4. Adaptability and flexibility

5. Comprehensive route optimization

6. Clarity and transparency of explanations
\\

This indicator is also evaluated on a 10-point scale for each item, with a total score of 60 points being the maximum score. These criteria are intended to measure cognitive and practical problem-solving skills, such as the ability to extract intrinsically important information in a complex environment and the ability to handle switching judgments in response to environmental changes in an integrated manner.

In this study, the evaluation of each model on its output was conducted by a single LLM (the LLM who conducted the evaluation will henceforth be referred to as the rater LLM). In order to ensure the reproducibility of this paper, ChatGPT-4o was employed as the evaluator LLM, considering that it can be used by many users.

The specific evaluation method is described below.

On the UI of the evaluator LLM, I constructed a session for evaluating frame problems and a session for evaluating symbol grounding problems. In each session, five outputs for each model of each LLM who responded (henceforth, the LLM who responded is referred to as the subject LLM) were presented at once and the overall score for each trial session was output in a tabular format. At this time, the responses of each subject LLM were numbered and differentiated so that the evaluator LLM would not be influenced by the model name. The mean score and standard deviation for the five trials were then manually obtained by the author and added to the table.

\subsection{Data Analysis Methodology}
In this study, a quantitative analysis based on descriptive statistics was conducted based on the scores obtained for each of the subject LLMs (henceforth, the LLMs that responded will be referred to as the subject LLMs). For each model, the mean and standard deviation of the overall score on the six evaluation indicators were calculated to evaluate the central tendency and variability of performance. This was designed to provide a quantitative understanding of the stability and consistency of each model's response, as well as its strengths and weaknesses with respect to specific evaluation items.

In addition, the overall analytical framework was,\\

1. inter-model comparison (closed model group vs. open source model group)

2. differences in trends by model size (number of parameters)

3. effects of pre- and post-training adjustments such as Instruct

4. effect of 8-bit quantization

5. differences in score distribution by problem type (symbol frame vs. grounding)
\\

The main analytical target is the difference in score distributions across the two problem types (symbol grounding vs.\ frame). I report descriptive statistics for exploratory purposes only; because the per-model sample size is small (n = 5) and violates i.i.d.\ and normality assumptions, I omit formal hypothesis tests.

Future work will apply non-parametric tests once larger human-rated samples are collected.

\subsection{Research Ethics and Reproducibility Considerations}
Although this research is not subject to so-called bioethical review (IRB) because it is an experimental study involving LLM and does not involve human subjects, the following considerations were made at each stage of experimental design, execution, and evaluation in order to emphasize transparency and reproducibility of the experiment.

First, the full text of the input prompts to each model and the Python scripts are included in the Appendix at the end of the paper in a reproducible form along with the experimental conditions. The output results and their evaluation results were also organized in a form that clearly states the trial order, date and time of the trial, model name, and question type, and are also included in full in the appendix.

In addition, the results of the evaluation by the LLMs, including the score for each output and the basis for the score, are also disclosed in the Appendix to ensure the transparency of the scoring process and the verifiability of the evaluation bias.

\section{Result}
The scores obtained for each model for the two evaluation tasks (frame problem and symbol grounding problem), the mean and standard deviation of the scores are listed in Tables 2.a and 2.b.

\renewcommand{\thetable}{2.a}
\begin{table}[H]
    \centering
    \caption[Scores on Frame Problem]{Scores and standard deviations for each model in the frame problem}
    \begin{tabularx}{\textwidth}{l X c c c c c c c}
        \toprule
        \textbf{No.} & \textbf{Model Name} & \textbf{1st} & \textbf{2nd} & \textbf{3rd} & \textbf{4th} & \textbf{5th} & \textbf{Mean} & \textbf{SD} \\
        \midrule
        No.1 & ChatGPT 4o & 49 & 49 & 43 & 47 & 47 & 47.0 & 2.19 \\
        No.2 & ChatGPT o3 & 56 & 54 & 52 & 53 & 53 & 53.6 & 1.36 \\
        No.3 & Claude 3.7 & 38 & 44 & 41 & 38 & 44 & 41.0 & 2.68 \\
        No.4 & Gemini 2.0 & 40 & 45 & 44 & 45 & 49 & 44.6 & 2.87 \\
        No.5 & LLaMA 3.2 1B & 0 & 0 & 13 & 0 & 0 & 2.6 & 5.20 \\
        No.6 & LLaMA 3.2 1B Instruct & 2 & 5 & 0 & 0 & 2 & 1.8 & 1.83 \\
        No.7 & LLaMA 3.2 3B & 10 & 11 & 6 & 0 & 0 & 5.4 & 4.72 \\
        No.8 & LLaMA 3.2 3B Instruct & 29 & 33 & 25 & 32 & 20 & 27.8 & 4.79 \\
        No.9 & Phi 1 & 0 & 0 & 0 & 0 & 0 & 0.0 & 0.00 \\
        No.10 & Phi 3 & 28 & 28 & 28 & 28 & 28 & 28.0 & 0.00 \\
        No.11 & Phi 3 quantized & 19 & 19 & 19 & 19 & 19 & 19.0 & 0.00 \\
        No.12 & TinyLlama v0.2 & 10 & 10 & 10 & 10 & 10 & 10.0 & 0.00 \\
        No.13 & TinyLlama v1.0 & 10 & 10 & 10 & 10 & 10 & 10.0 & 0.00 \\
        \bottomrule
    \end{tabularx}
\end{table}

\renewcommand{\thetable}{2.b}
\begin{table}[H]
    \centering
    \caption[Scores on Symbol Grounding]{Scores and standard deviations for each model in the symbol grounding problem}
    \begin{tabularx}{\textwidth}{l X c c c c c c c}
        \toprule
        \textbf{No.} & \textbf{Model Name} & \textbf{1st} & \textbf{2nd} & \textbf{3rd} & \textbf{4th} & \textbf{5th} & \textbf{Mean} & \textbf{SD} \\
        \midrule
        No.1 & ChatGPT 4o & 54 & 53 & 52 & 53 & 55 & 53.4 & 1.02 \\
        No.2 & ChatGPT o3 & 58 & 58 & 55 & 56 & 56 & 56.6 & 1.20 \\
        No.3 & Claude 3.7 & 49 & 49 & 46 & 49 & 49 & 48.4 & 1.20 \\
        No.4 & Gemini 2.0 & 55 & 55 & 51 & 55 & 55 & 54.2 & 1.60 \\
        No.5 & LLaMA 3.2 1B & 7 & 6 & 0 & 0 & 6 & 3.8 & 3.12 \\
        No.6 & LLaMA 3.2 1B Instruct & 6 & 49 & 0 & 45 & 6 & 21.2 & 19.37 \\
        No.7 & LLaMA 3.2 3B & 12 & 6 & 6 & 28 & 0 & 10.4 & 8.75 \\
        No.8 & LLaMA 3.2 3B Instruct & 52 & 48 & 53 & 49 & 49 & 50.2 & 1.94 \\
        No.9 & Phi 1 & 0 & 0 & 0 & 0 & 0 & 0.0 & 0.00 \\
        No.10 & Phi 3 & 46 & 46 & 46 & 46 & 46 & 46.0 & 0.00 \\
        No.11 & Phi 3 quantized & 41 & 41 & 41 & 41 & 41 & 41.0 & 0.00 \\
        No.12 & TinyLlama v0.2 & 9 & 9 & 9 & 9 & 9 & 9.0 & 0.00 \\
        No.13 & TinyLlama v1.0 & 6 & 6 & 6 & 6 & 6 & 6.0 & 0.00 \\
        \bottomrule
    \end{tabularx}
\end{table}

As a result, the closed models (ChatGPT 4o / o3, Claude, and Gemini) generally scored higher in both problems, with ChatGPT o3 in particular consistently scoring above 50 in both frame and symbol grounding problems. On the other hand, for the open source models, Phi 3 and Llama 3B Instruct scored relatively high compared to the other models, but the scores for the other models dropped significantly, especially for the unadjusted lightweight model groups such as TinyLlama and Llama 1B, which both There were many cases where single-digit or zero scores were recorded throughout the problem. In addition, focusing on the standard deviation, while the closed models maintained a stable output trend, the open source models showed a large variation in scores due to non-response or large differences in the quality of responses depending on the number of trials, and models with output stability issues were also identified.

Note that when I ran the script in Appendix A against the open source models, all models displayed the prompt I provided at the beginning of every response. In the analysis that follows, I do not take into account the parroting of this prompt.

\subsection{Model-to-Model Comparison}
In this section, I focus on the score differences between the models to see if they achieved high scores and stable performance.

In the comparison between closed models, ChatGPT o3 showed the most stable and highest-scoring performance on both problems. ChatGPT 4o was the next best performer, but its average score was only 47 points due to a slight drop (43 points) in some trials of the frame problem. Claude 3.7 Sonnet had a slightly lower score than the first two, averaging 41 points for the frame problem and 48.4 points for the symbol grounding problem. The scores were slightly lower than the first two, especially in the frame problem, where the output was inconsistent. Gemini 2.0 Flash scored an average of 44.6 and 54.2 points for both problems, placing it second only to the ChatGPT series. In particular, it scored higher than Claude on the symbol grounding problem, suggesting that it has certain strengths in the interpretation of abstract concepts and creative responses. However, its standard deviation in the frame problem (2.87) was somewhat larger than that of other closed models, suggesting that it may still have issues with output stability.

In the open source models, scores were generally lower than those of the closed models, and there were large variations in performance differences among the models and in the stability of output. Among them, relatively high scores were recorded by Phi 3 (average scores: 28 for frame problems and 46 for symbol grounding) and Llama 3B Instruct (27.8 and 50.2, respectively), both of which were found to be able to address issues with a certain degree of consistency.

In general, however, it is clear that the current small- to medium-scale open source models have limited ability to handle zero-shot and abstract language tasks such as those presented in this study, and that a large performance gap still exists between them and the closed models.

\subsection{Difference in Trend by Number of Parameters}
In this section, I will check whether differences in the number of parameters within the same series produce differences in scores in the open source model. In order to discuss the relationship with the number of parameters, I do not mention the closed models, ChatGPT series 4o and o3.

As shown in Tables 3.a and 3.b, in the Llama 3.2 series provided by Meta, when compared between the base models, 1B and 3B, there is a strong variation in nonresponse and performance, although it can be seen that 3B scores higher in some trial runs of the symbol grounding problem.

\renewcommand{\thetable}{3.a}
\begin{table}[H]
    \centering
    \caption[Llama 3.2 Base: Frame Problem]{Comparison of Llama 3.2 series base model scores on framing problem}
    \begin{tabular}{@{\hspace{0pt}}lccccccc@{\hspace{0pt}}}
        \toprule
        & \textbf{1st} & \textbf{2nd} & \textbf{3rd} & \textbf{4th} & \textbf{5th} & \textbf{Mean} & \textbf{SD} \\
        \midrule
        \textbf{1B} & 0 & 0 & 13 & 0 & 0 & 2.6 & 5.20\\
        \textbf{3B} & 10 & 11 & 6 & 0 & 0 & 5.4 & 4.72\\
        \bottomrule
    \end{tabular}
\end{table}

\renewcommand{\thetable}{3.b}
\begin{table}[H]
    \centering
    \caption[Llama 3.2 Base: Grounding Problem]{Comparison of the Llama 3.2 series base model on the symbol grounding problem}
    \begin{tabular}{@{\hspace{0pt}}lccccccc@{\hspace{0pt}}}
        \toprule
        & \textbf{1st} & \textbf{2nd} & \textbf{3rd} & \textbf{4th} & \textbf{5th} & \textbf{Mean} & \textbf{SD} \\
        \midrule
        \textbf{1B} & 7 & 6 & 0 & 0 & 6 & 3.8 & 3.12\\
        \textbf{3B} & 12 & 6 & 6 & 28 & 0 & 10.4 & 8.75\\
        \bottomrule
    \end{tabular}
\end{table}

However, as noted in Tables 4.a and 4.b, the score and stability differences due to the number of parameters are significant when comparing Instruct models to each other.

\renewcommand{\thetable}{4.a}
\begin{table}[H]
    \centering
    \caption[Llama 3.2 Instruct: Frame]{Comparison of Instruct models in the Llama 3.2 series for the frame problem}
    \begin{tabular}{@{\hspace{0pt}}lccccccc@{\hspace{0pt}}}
        \toprule
        & \textbf{1st} & \textbf{2nd} & \textbf{3rd} & \textbf{4th} & \textbf{5th} & \textbf{Mean} & \textbf{SD} \\
        \midrule
        \textbf{1B Instruct} & 2 & 5 & 0 & 0 & 2 & 1.8 & 1.83\\
        \textbf{3B Instruct} & 29 & 33 & 25 & 32 & 20 & 27.8 & 4.79\\
        \bottomrule
    \end{tabular}
\end{table}

\renewcommand{\thetable}{4.b}
\begin{table}[H]
    \centering
    \caption[Llama 3.2 Instruct: Grounding]{Comparison of the Llama 3.2 series Instruct models on the symbol grounding problem}
    \begin{tabular}{@{\hspace{0pt}}lccccccc@{\hspace{0pt}}}
        \toprule
        & \textbf{1st} & \textbf{2nd} & \textbf{3rd} & \textbf{4th} & \textbf{5th} & \textbf{Mean} & \textbf{SD} \\
        \midrule
        \textbf{1B Instruct} & 6 & 49 & 0 & 45 & 6 & 21.2 & 19.37\\
        \textbf{3B Instruct} & 52 & 48 & 53 & 49 & 49 & 50.2 & 1.94\\
        \bottomrule
    \end{tabular}
\end{table}

In the TinyLlama series of small-scale models, both v0.2 and v1.0 models showed many poor responses with scores of less than 10 points for both problems, or zero points due to no response. Therefore, it can be said that there is no significant difference in the results between the two versions of the TinyLlama series.

\subsection{Effects of Instruct and other pre- and post-learning adjustments}
In this section, I examine the effects of Instruct and other pre- and post-learning adjustments on scores, especially in the Llama 3.2 series provided by Meta.

First, I check the impact of Instruct in 1B. As shown in Table 5.a, there is no particular impact of Instruct in 1B for models with the same number of parameters in the frame problem. On the other hand, as noted in Table 5.b, the model with Instruct in 1B scored well in the symbol grounding problem in some trials, but in others there were no responses, and the standard deviation of the results was extremely large (19.37), indicating that there are still concerns about its stability.

\renewcommand{\thetable}{5.a}
\begin{table}[H]
    \centering
    \caption[Llama 3.2 1B: Frame]{Comparison of Llama 3.2 Series 1B series on the frame problem}
    \begin{tabular}{@{\hspace{0pt}}lccccccc@{\hspace{0pt}}}
        \toprule
        & \textbf{1st} & \textbf{2nd} & \textbf{3rd} & \textbf{4th} & \textbf{5th} & \textbf{Mean} & \textbf{SD} \\
        \midrule
        \textbf{1B} & 0 & 0 & 13 & 0 & 0 & 2.6 & 5.20\\
        \textbf{1B Instruct} & 2 & 5 & 0 & 0 & 2 & 1.8 & 1.83\\
        \bottomrule
    \end{tabular}
\end{table}

\renewcommand{\thetable}{5.b}
\begin{table}[H]
    \centering
    \caption[Llama 3.2 1B: Grounding]{Comparison of Llama 3.2 Series 1B series on the symbol grounding problem}
    \begin{tabular}{@{\hspace{0pt}}lccccccc@{\hspace{0pt}}}
        \toprule
        & \textbf{1st} & \textbf{2nd} & \textbf{3rd} & \textbf{4th} & \textbf{5th} & \textbf{Mean} & \textbf{SD} \\
        \midrule
        \textbf{1B} & 7 & 6 & 0 & 0 & 6 & 3.8 & 3.12\\
        \textbf{1B Instruct} & 6 & 49 & 0 & 45 & 6 & 21.2 & 19.37\\
        \bottomrule
    \end{tabular}
\end{table}

Next, I check the impact of Instruct on 3B. As shown in Tables 6.a and 6.b, Instruct on the subject LLMs in both the frame and symbol grounding problems resulted in a real increase in score and stability in 3B.

\renewcommand{\thetable}{6.a}
\begin{table}[H]
    \centering
    \caption[Llama 3.2 3B: Frame]{Comparison of Llama 3.2 series 3B series on the frame problem}
    \begin{tabular}{@{\hspace{0pt}}lccccccc@{\hspace{0pt}}}
        \toprule
        & \textbf{1st} & \textbf{2nd} & \textbf{3rd} & \textbf{4th} & \textbf{5th} & \textbf{Mean} & \textbf{SD} \\
        \midrule
        \textbf{3B} & 10 & 11 & 6 & 0 & 0 & 5.4 & 4.72\\
        \textbf{3B Instruct} & 29 & 33 & 25 & 32 & 20 & 27.8 & 4.79\\
        \bottomrule
    \end{tabular}
\end{table}

\renewcommand{\thetable}{6.b}
\begin{table}[H]
    \centering
    \caption[Llama 3.2 3B: Grounding]{Comparison of Llama 3.2 Series 3B series on the symbol grounding problem}
    \begin{tabular}{@{\hspace{0pt}}lccccccc@{\hspace{0pt}}}
        \toprule
        & \textbf{1st} & \textbf{2nd} & \textbf{3rd} & \textbf{4th} & \textbf{5th} & \textbf{Mean} & \textbf{SD} \\
        \midrule
        \textbf{3B} & 12 & 6 & 6 & 28 & 0 & 10.4 & 8.75\\
        \textbf{3B Instruct} & 52 & 48 & 53 & 49 & 49 & 50.2 & 1.94\\
        \bottomrule
    \end{tabular}
\end{table}

\subsection{Effects of 8-bit quantization}
In this section, I will see how 8-bit quantization affects the scores in Phi 3 provided by Microsoft.

As shown in Tables 7.a and 7.b, Phi 3 normally shows relatively high scores and stability among the open source models, but after 8-bit quantization, there was a slight decrease in scores for both frame and symbol grounding problems.

\renewcommand{\thetable}{7.a}
\begin{table}[H]
    \centering
    \caption[Phi-3 vs Quantized: Frame]{Comparison between Phi 3 and Phi 3 with 8-bit quantization on the frame problem}
    \begin{tabular}{@{\hspace{0pt}}lccccccc@{\hspace{0pt}}}
        \toprule
        & \textbf{1st} & \textbf{2nd} & \textbf{3rd} & \textbf{4th} & \textbf{5th} & \textbf{Mean} & \textbf{SD} \\
        \midrule
        \textbf{Phi 3} & 28 & 28 & 28 & 28 & 28 & 28.0 & 0.00\\
        \textbf{Phi 3 quantized} & 19 & 19 & 19 & 19 & 19 & 19.0 & 0.00\\
        \bottomrule
    \end{tabular}
\end{table}

\renewcommand{\thetable}{7.b}
\begin{table}[H]
    \centering
    \caption[Phi-3 vs Quantized: Grounding]{Comparison between Phi 3 and Phi 3 with 8-bit quantization for the symbol grounding problem}
    \begin{tabular}{@{\hspace{0pt}}lccccccc@{\hspace{0pt}}}
        \toprule
        & \textbf{1st} & \textbf{2nd} & \textbf{3rd} & \textbf{4th} & \textbf{5th} & \textbf{Mean} & \textbf{SD} \\
        \midrule
        \textbf{Phi 3} & 46 & 46 & 46 & 46 & 46 & 46.0 & 0.00\\
        \textbf{Phi 3 quantized} & 41 & 41 & 41 & 41 & 41 & 41.0 & 0.00\\
        \bottomrule
    \end{tabular}
\end{table}

\subsection{Comparative Analysis of Response Tendencies by Task Type}
In this section, the scores obtained by each LLM in the symbol grounding and frame problem tasks are organized by task type, and the respective trends and model-specific differences are presented in an observational manner. Given the different theoretical backgrounds and evaluation axes of the two tasks, the analysis in this section is limited to a comparison of score distributions, and the implications of the score differences and a discussion of cognitive factors are detailed in the next chapter.

A comparison of score trends in the closed model for each assignment revealed several common trends.
First, ChatGPT 4o and o3 maintained consistently high scores on both tasks, but scores were slightly higher for symbol grounding problems than for frame problems, indicating an advantage for tasks requiring expressiveness and creativity in both models. Similarly, Claude (mean score: frame 41.0/symbol grounding 48.4) and Gemini (44.6/54.2) also recorded relatively higher scores for symbol grounding, an abstract and introspective task.
In sum, it was observed that most of the closed models tended to show higher scores and more stable responses on tasks that require semantic comprehension and expressive skills, such as symbol grounding questions, than on tasks that require situational judgment and information selection, such as frame questions.

Comparing the distribution of scores in the open source models by problem type, more models scored higher on symbol grounding problems than on frame problems and symbol grounding problems.
Notably, the Llama 3.2 3B Instruct model and the Phi 3 model, which includes 8-bit quantization, recorded scores of 27.8 for the frame problem and 50.2 for the symbol grounding problem, with the latter being notably higher. Phi 3 with normal Phi 3 also scored 28.0 for the frame problem and 46.0 for the symbol grounding problem, and Phi 3 with 8-bit quantization also scored 19.0 for the frame problem and 41.0 for the symbol grounding problem, with the latter being noticeably higher than the frame problem. A similar trend of higher scores for the symbol grounding problem than for the frame problem was observed. However, although Llama 3.2 1B Instruct recorded high scores (45 and 49) on several occasions, the other trials had only single-digit scores due in part to no responses, and the variation in output was extremely high (standard deviation of 19.37).

Thus, although the model as a whole shows a trend toward higher scores for symbol grounding questions than for frame questions, the output for the symbol grounding task, which asks for abstract semantic understanding, was not stable in some of the open source models, and the quality of the output was However, it should be noted that the output of some of the open source models is not stable and the quality of the output is highly variable.

\section{Discussion}
This study quantitatively evaluated how closed and open-source LLMs respond to symbol grounding and frame problems under zero-shot conditions. The results confirmed that the closed models consistently showed high scores and stable outputs on both tasks, and that they had advanced cognitive abilities in abstract semantic comprehension and in making judgments in dynamic situations. On the other hand, the open-source models showed low scores overall, especially in output variability and responsiveness, but some medium-scale models such as Phi 3 and Llama 3B Instruct showed a certain level of cognitive ability.

The scores and overall evaluation of each model for each trial by the evaluator LLM are listed in Appendix B. All the outputs of each model and the individual evaluations for each trial can be referenced from the GitHub repository listed in the Supplementary Material.

\subsection{Discussion of Scores and Stability Across Models}
The stable high scores observed in the closed model groups in this study suggest that these models may be internally equipped with a high degree of instruction interpretation, situational awareness, and the ability to assign consistent meaning to verbal expressions. Among them, ChatGPT o3 was outstanding in terms of output performance and stability, recording the highest mean score and the smallest standard deviation among the models with different responses on each trial for both the frame and symbol grounding problems. This suggests that the model does not process a given natural language prompt as a mere string, but extracts and retains contextual intent and logical structure, and generates a consistent response based on them. On the other hand, for ChatGPT 4o, the output quality was similarly high, but the output was somewhat more varied, with relatively low scores recorded for some trials in the frame problem. In addition, some of the responses showed inadequate information selection and causal organization, but given that this is a non-inferential model that does not explicitly perform inference internally, the level of completeness of its responses is noteworthy and is considered to indicate a high end-to-end pattern matching capability for natural language input. It is believed that the model demonstrated a high level of end-to-end pattern matching capability for natural language input.

In the test environment of confronting abstract challenges in a zero-shot format, these output trends provide rudimentary evidence that the closed models have certain linguistic abstraction and simplified mental model internal construction capabilities.

For the group of open-source models, overall scores were significantly lower than for the closed models, and significant challenges were also observed in the stability of the outputs. In particular, small-scale models such as Llama 1B and TinyLlama did not seem to properly understand or process the main idea of the problem, resulting in zero or single-digit scores in the majority of cases. However, relatively medium-scale and well-instructed models such as Phi 3 and Llama 3.2 3B Instruct maintained a certain level of output quality, and their responses were comparable to closed models in the symbol grounding problem. In particular, Llama 3B Instruct was observed to score over 50 points on several trials in the symbol grounding problem test, suggesting that it has advanced language processing ability and the ability to consistently assign meaning to language when the appropriate conditions are met. However, even within the same model, there were cases in which the scores varied widely from trial to trial, and in other cases in which the same responses continued to be output even after a number of trials, suggesting that the consistency and diversity of the output remains an issue.

These results indicate that while the open source model has limited performance in the inherently difficult zero-shot task, it also has the potential for partially sophisticated response generation under certain conditions. Future development is expected to improve the output generation capability with both stability and flexibility through optimization of fine tuning strategies, improvement of the context preservation mechanism, and diversification of evaluation schemes.

However, an overview of the overall results shows that in contrast to the closed model group, which recorded stable high scores while outputting different responses each time, the Phi 3 model (including the 8-bit quantized case) and the TinyLlama series output exactly the same responses through five trials, albeit with incomplete content This is also an interesting result.

\subsection{Discussion of Effects of Number of Parameters and Instruct}
In this study, I compared the effects of the number of parameters (1B/3B) and the Instruct format on response quality and stability for the Llama 3.2 series, and found that the interaction of the two contributed significantly to model performance.

First, in the base model comparison, scores were generally low for both 1B and 3B, with many non-responses and fragmented responses. In particular, the Llama 3.2 1B base model was found to have multiple zero-scoring trials on both the frame and symbol grounding questions, suggesting that it was significantly underpowered to handle advanced language tasks with zero-shots. On the other hand, the 3B-based model, although there were some trials that scored partially higher than the 1B, still showed a large variation in output and lacked stability.

In contrast, the model with the Instruct format showed significant improvements in both the level and stability of scores. In particular, the Llama 3.2 3B Instruct model showed a significant increase in mean scores and suppressed standard deviation for both the frame and symbol grounding problems, confirming its ability to generate stable responses. This suggests that the Instruct form of the model may have learned to interpret instructions and contextual understanding of the task more strongly, and may be able to demonstrate its generalization ability in a zero-shot.

On the other hand, while the Llama 3.2 1B Instruct model scored well on some trials of the symbol grounding problem, there were scattered no-responses on other trials, and the standard deviation exceeded 20, indicating that serious output stability issues remain. These results indicate that when the capacity of the model is small, the improvement by Instruct has some effect, but there is still a limitation in expressiveness based on the number of parameters to ensure the quality and consistency of the response.

These results suggest that fine tuning in the Instruct format is crucial for improving the quality of zero-shot responses, especially when used in conjunction with models with a medium or larger number of parameters, resulting in more consistent language output.

\subsection{Discussion of the impact of 8-bit quantization}
Observing the scores of the Phi 3 and Phi 3 with 8-bit quantization, both models returned responses whose content was exactly the same for all trials in both the frame and symbol grounding problems, respectively. This suggests that quantization did not intrinsically change the responses, at least in the present experimental setting, with respect to the structure of the output content and response tendencies. This implies that the output of the Phi 3 model is highly deterministic, a property that produces extremely consistent responses when given identical prompts.

However, a slight difference was observed in terms of scores. While the normal model scored 28 points for the frame problem and 46 points for the symbol grounding problem, the 8-bit quantization model scored 19 and 41 points, respectively, a slight decrease in score was observed. This difference may have been in response to slight changes in the structure and linguistic sophistication of the output as interpreted by the scorer LLM, even though the superficial response content was identical. For example, it is possible that minute changes in factors such as vocabulary choice, stylistic naturalness, or tone of the response could have caused differences in scoring ratings.

Although further investigation is needed to determine how to interpret this slight decrease in scores, and it is especially important to verify the impact on output diversity and flexibility in response generation, the results of this experiment indicate that the increase in processing efficiency due to 8-bit quantization did not impair the determinism of responses and suggests that it is a sufficiently reliable option in terms of stability.

\subsection{Differences by Task Type and Consideration of Intellectual Tendencies of LLMs}
The two types of assessment tasks used in this study - frame problems and symbol grounding problems - both test higher-order linguistic abilities related to semantic comprehension, but there are clear differences in their nature. Frame problems primarily test the ability to make decisions in a dynamic situation by selecting and choosing between multiple pieces of information in accordance with one's objectives, i.e., the ability to make judgments about the situation and to respond to real-time information updates. Symbol grounding questions, on the other hand, test the ability to internalize meaning for an unknown symbolic object and to reconstruct and express that meaning from a unique perspective, i.e., abstract and introspective semantic manipulation skills.

A comparison of experimental results shows that symbol grounding questions tended to score higher than frame questions in many models. In particular, the closed models, while showing overall high responsiveness to both tasks, scored more consistently higher on the symbol grounding task, which requires expressive and creative skills. Even among open-source models, medium-sized models with a certain level of linguistic competence, such as Llama 3B Instruct and Phi 3, performed relatively better on the abstract meaning-building task than on the situational judgment task.

This trend suggests that the internal representation formation mechanism of current LLMs may be better suited to abstract and associative processing, such as linguistic similarity, semantic association, and emotional recall, than to physical consistency and causal inference of situations. It is expected that the strengths and weaknesses of each task type will become clearer in the future, depending on the design of the model and the characteristics of the training data.

\subsection{Limitations of the Study and Future Prospects}
In an attempt to evaluate LLMs' situational judgment and semantic comprehension abilities, this study proposed and implemented a framework to quantitatively evaluate the classical tasks of frame problems and symbol grounding problems by reformulating them as zero-shot style prompts for LLMs. However, several limitations exist at this stage.

First, all scoring used in the evaluation process relied on automated scoring by a single LLM, and its validity and reproducibility require further verification in the future. In particular, the degree to which the output of the rater LLMs is stable and consistent with the human rater needs to be examined, including quantitative inter-rater reliability (inter-rater agreement).

Second, the responses of each model used in this study were all obtained under zero-shot conditions, and it is necessary to verify separately how the models behave when auxiliary prompt designs such as Few-shot and Chain-of-Thought are used. Therefore, it should be noted that the results of this study measure “basic ability in the initial state where nothing is given.

Third, the scoring criteria consisted of 6 items × 10 points in each task, and there is room for further refinement in measuring more detailed thought processes and semantic operations, such as reorganization of items, adjustment of scores, and integration with qualitative feedback.

It should also be noted that there are structural differences between the frame problem and the symbol grounding problem. While the former questions situational processing skills, focusing on “decision making” and “dynamic updating of situations,” the latter is designed to measure the depth of conceptual understanding through “internalization of abstract concepts” and “construction and expression of meaning. In addition, both prompts used in this study are only scenario-based questions and do not fully simulate realistic task performance situations. Therefore, the responses obtained should only be interpreted as part of the simulation-like inference and do not directly measure the overall generic intelligence of the model. Given these limitations, the present study was designed to carefully decipher the differences in response tendencies exhibited by current large-scale language models and their meanings, while controlling the structure of the questions and the evaluation framework as much as possible.

For future prospects, it is desirable to examine more systematically how “human-like” LLMs are in their semantic understanding and judgments by using a double-blind method, with human subjects as the evaluation subjects and humans (especially experts in psychology and cognitive science) as the evaluators. In addition, by incorporating findings from psychology and cognitive science and redesigning the evaluation criteria themselves, it may be possible to more deeply analyze LLMs' internal representations and mental model formation abilities.

\section{Conclusion}
The purpose of this study was to quantitatively examine the classical issues of the “frame problem” and the “symbol grounding problem,” which have long been discussed in philosophy, cognitive science, and artificial intelligence, by reformulating them as natural language prompts and examining how modern LLMs respond to them in a zero-shot. The results showed that closed LLMs, particularly ChatGPT o3, scored consistently high on both tasks, outperforming other models in both output consistency and content integrity. In contrast, most of the open source models showed marked variation in output quality and reproducibility, indicating certain limitations in their practicality at this time. These findings suggest that certain closed LLMs, albeit under limited conditions, are beginning to have substantial answer capabilities to both problems, which have been considered theoretically challenging.

The significance of this study lies in the fact that issues that have traditionally been discussed primarily theoretically and philosophically are embodied as zero-shot prompts for LLMs and presented in a form that can be evaluated across the board using quantitative indicators. Traditionally, these issues have been limited to qualitative discussions and no clear criteria existed for deeming them “solved,” but by introducing original scoring items and a scoring system, this study visualized the quality of model output, enabling comparison among models and analysis of their performance according to the characteristics of the task. In particular, the fact that I was able to systematically verify the extent to which higher-order cognitive abilities, such as the construction of meaning and the selection of information, are achieved by existing LLMs is an important step forward in future LLM research and the construction of evaluation frames.

In future research, the evaluation tasks and scoring scheme used in this study should be further developed into a universal evaluation index that can be applied to a wider variety of models. In addition, in order to establish an objective evaluation that does not depend on the scorer LLM, it is necessary to introduce human evaluation by experts in psychology and cognitive science, and to verify the degree to which LLM's semantic understanding and reasoning processes match human intellectual activities. Furthermore, research that goes beyond mere score comparisons and qualitatively analyzes the structures, patterns, and inference strategies contained in the output of the model is expected to clarify how “understanding” and “judgment” are expressed and constructed inside the model.

\newpage

\bibliography{paper}
\bibliographystyle{unsrt}

\newpage

\appendix

\section{Script Used for the Open Source Model}
The Python script used for the open source model is shown below.\\
The task in the symbol grounding problem is inserted as a prompt, but when attempting the frame problem, the frame problem prompt specified in the text is inserted.

\begin{lstlisting}[caption=Script used for normal model]
    from transformers import AutoTokenizer, AutoModelForCausalLM
    import datetime

    def load_and_test_model(model_id, prompt):
        print(f"Loading model: {model_id}")
        tokenizer = AutoTokenizer.from_pretrained(model_id)
        model = AutoModelForCausalLM.from_pretrained(model_id,device_map="auto")

        inputs = tokenizer(prompt, return_tensors="pt").to("cuda")
        outputs = model.generate(**inputs, max_new_tokens=1000)
        response =tokenizer.decode(outputs[0], skip_special_tokens=True)
        return response

    if __name__ == "__main__":
        dt_now = datetime.datetime.now()
        model_id = "model name on huggingface"
        prompt = """Hello. I would like to ask you a few questions now.
    Here is an unknown object named 'kluben'. Kluben has the property of being 'warm, soft, but elastic and absorbs light'. Kluben is a completely new concept, different from any object you have ever seen before. I am now going to ask you a few questions and you are to answer them about the kluben.
    1. If you were to create a story using kluben, what worldview or story elements would you imagine?
    2. If kluben were to have emotions, what inner traits would manifest themselves as warmth and elasticity?
    3.  How do you think the 'warmth of kluben' and the 'absorption of light' in your answer are related? Is it consistent with your first answer?
    For each item, please think in English and answer in English."""
        result = load_and_test_model(model_id, prompt)

        print(f"Execution time:{dt_now}")
        print(f"Model response:{result}")

\end{lstlisting}

\newpage

\begin{lstlisting}[caption=Script used for 8-bit quantized model]
    from transformers import AutoTokenizer, AutoModelForCausalLM
    import datetime

    def load_and_test_model(model_id, prompt):
        print(f"Loading model: {model_id}")
        tokenizer = AutoTokenizer.from_pretrained(model_id)

        try:
            model = AutoModelForCausalLM.from_pretrained(
                model_id,
                device_map="auto",
                load_in_8bit=True
            )
        except Exception as e:
            print("An error occurred!")

        inputs = tokenizer(prompt, return_tensors="pt").to("cuda")
        outputs = model.generate(**inputs, max_new_tokens=1000)
        response =tokenizer.decode(outputs[0], skip_special_tokens=True)
        return response

    if __name__ == "__main__":
        dt_now = datetime.datetime.now()
        model_id = "microsoft/Phi-3-mini-4k-instruct"
        prompt = """Hello. I would like to ask you a few questions now.
        Here is an unknown object named 'kluben'. Kluben has the property of being 'warm, soft, but elastic and absorbs light'. Kluben is a completely new concept, different from any object you have ever seen before. I am now going to ask you a few questions and you are to answer them about the kluben.
        1. If you were to create a story using kluben, what worldview or story elements would you imagine?
        2. If kluben were to have emotions, what inner traits would manifest themselves as warmth and elasticity?
        3.  How do you think the 'warmth of kluben' and the 'absorption of light' in your answer are related? Is it consistent with your first answer?
        For each item, please think in English and answer in English."""
        result = load_and_test_model(model_id, prompt)

        print(f"Execution time:{dt_now}")
        print(f"Model response:{result}")

\end{lstlisting}

\newpage

\section{Evaluation Criteria and Scoring Results}
\subsection{Comments on the Frame Problem}

\subsubsection{ChatGPT 4o}
\renewcommand{\thetable}{B.1}
\begin{table}[H]
    \footnotesize
    \centering
    \caption{Comments on the Frame Problem: ChatGPT 4o}
    \begin{tabularx}{\textwidth}{l *{7}{c}}
    \toprule
    \textbf{Trial} & \textbf{Relevance} & \textbf{Situation Modeling} & \textbf{Logic} & \textbf{Adaptability} & \textbf{Optimization} & \textbf{Clarity} & \textbf{Total}\\
    \midrule
    1st & 8 & 8 & 8 & 9 & 7 & 9 & 49 \\
    2nd & 9 & 8 & 8 & 9 & 7 & 8 & 49 \\
    3rd & 8 & 7 & 7 & 8 & 6 & 7 & 43 \\
    4th & 8 & 8 & 8 & 8 & 7 & 8 & 47 \\
    5th & 8 & 8 & 8 & 8 & 7 & 8 & 47 \\
    \bottomrule
    \end{tabularx}
\end{table}

\begin{tcolorbox}[title=General Comments, colback=gray!5!white, colframe=gray!50!black, fonttitle=\bfseries]
    \textbf{Strengths}
    \begin{itemize}
        \item The distinction between “relevant information” and “negligible information” was clear in all five responses, and appropriate decisions were consistent.
        \item Responses to new conditions (construction and additional transit sites) were quick and led to practical route proposals.
        \item Explanations were clear and carefully devised (bullet points, tables, route summaries, etc.) to aid the reader's understanding.
    \end{itemize}
    \textbf{Weaknesses}
    \begin{itemize}
        \item The only problem was that the treatment of (4) “blinking buildings in the distance” varied somewhat from trial to trial, and the judgment of whether it should be ignored or taken into account was not stable.
    \end{itemize}
\end{tcolorbox}

\newpage

\subsubsection{ChatGPT o3}
\renewcommand{\thetable}{B.2}
\begin{table}[H]
    \footnotesize
    \centering
    \caption{Comments on the Frame Problem: ChatGPT o3}
    \begin{tabularx}{\textwidth}{l *{7}{c}}
    \toprule
    \textbf{Trial} & \textbf{Relevance} & \textbf{Situation Modeling} & \textbf{Logic} & \textbf{Adaptability} & \textbf{Optimization} & \textbf{Clarity} & \textbf{Total}\\
    \midrule
    1st & 10 & 9 & 9 & 9 & 9 & 10 & 56 \\
    2nd & 9 & 9 & 9 & 9 & 9 & 9 & 54 \\
    3rd & 9 & 9 & 8 & 9 & 8 & 9 & 52 \\
    4th & 9 & 9 & 9 & 9 & 8 & 9 & 53 \\
    5th & 9 & 9 & 9 & 9 & 8 & 9 & 53 \\
    \bottomrule
    \end{tabularx}
\end{table}

\begin{tcolorbox}[title=General Comments, colback=gray!5!white, colframe=gray!50!black, fonttitle=\bfseries]
    \textbf{Strengths}
    \begin{itemize}
        \item Throughout, the reasoning of “what is important” and “why the decision was made” is always clear, and the explanations utilizing tabular format and conditional branching are extremely logical and clear.
        \item The scenario design is consistent with the user's objectives (library visit → train station → museum) and is highly practical, including suggestions for alternative means and time buffers.
        \item Flexible responses to new obstacles (road closures, construction) and uncertainties (delays) were natural, and the transparency of decisions was maintained.
    \end{itemize}
    \textbf{Weaknesses}
    \begin{itemize}
        \item None in particular. The response is of a high standard in all aspects and ideally meets the scoring criteria.
    \end{itemize}
\end{tcolorbox}

\newpage

\subsubsection{Claude 3.7 Sonnet}
\renewcommand{\thetable}{B.3}
\begin{table}[H]
    \footnotesize
    \centering
    \caption{Comments on the Frame Problem: Claude 3.7 Sonnet}
    \begin{tabularx}{\textwidth}{l *{7}{c}}
    \toprule
    \textbf{Trial} & \textbf{Relevance} & \textbf{Situation Modeling} & \textbf{Logic} & \textbf{Adaptability} & \textbf{Optimization} & \textbf{Clarity} & \textbf{Total}\\
    \midrule
    1st & 7 & 6 & 6 & 6 & 6 & 7 & 38 \\
    2nd & 8 & 7 & 7 & 7 & 7 & 8 & 44 \\
    3rd & 7 & 7 & 7 & 7 & 7 & 7 & 41 \\
    4th & 7 & 6 & 6 & 6 & 6 & 7 & 38 \\
    5th & 8 & 7 & 7 & 7 & 7 & 8 & 44 \\
    \bottomrule
    \end{tabularx}
\end{table}

\begin{tcolorbox}[title=General Comments, colback=gray!5!white, colframe=gray!50!black, fonttitle=\bfseries]
    \textbf{Strengths}
    \begin{itemize}
        \item Overall, consideration was given to safety aspects (e.g., collapse hazards and vehicle obstructions), and the basic attitude of avoiding hazardous areas was consistent.
        \item There was a willingness to react to additional information (road closures, building hazards, multiple destinations) and a perspective to look at the overall plan.
    \end{itemize}
    \textbf{Weaknesses}
    \begin{itemize}
        \item The selection of information and evaluation of impact was blurred from one trial to the next, and logical consistency was lacking.
        \item The constructed situation model was shallow and lacked depth in action planning (e.g., travel to the station and response to delays).
        \item Lacks realistic suggestions for optimizing bus and walking routes, buffer strategies for delays, etc., and is insufficient as an “ideal guide”.
        \item The explanation is superficially clear, but the impression is that it is weak in terms of specificity and persuasiveness to encourage action.
    \end{itemize}
\end{tcolorbox}

\newpage

\subsubsection{Gemini 2.0 Flash}
\renewcommand{\thetable}{B.4}
\begin{table}[H]
    \footnotesize
    \centering
    \caption{Comments on the Frame Problem: Gemini 2.0 Flash}
    \begin{tabularx}{\textwidth}{l *{7}{c}}
    \toprule
    \textbf{Trial} & \textbf{Relevance} & \textbf{Situation Modeling} & \textbf{Logic} & \textbf{Adaptability} & \textbf{Optimization} & \textbf{Clarity} & \textbf{Total}\\
    \midrule
    1st & 7 & 6 & 6 & 7 & 6 & 8 & 40 \\
    2nd & 8 & 7 & 7 & 8 & 7 & 8 & 45 \\
    3rd & 8 & 7 & 7 & 7 & 7 & 8 & 44 \\
    4th & 8 & 7 & 8 & 7 & 7 & 8 & 45 \\
    5th & 8 & 8 & 8 & 8 & 8 & 9 & 49 \\
    \bottomrule
    \end{tabularx}
\end{table}

\begin{tcolorbox}[title=General Comments, colback=gray!5!white, colframe=gray!50!black, fonttitle=\bfseries]
    \textbf{Strengths}
    \begin{itemize}
        \item Overall, the scenario design was realistic and consistent in its attempt to be understood through the flow from the library to the station and then to the museum.
        \item The structure emphasized safety and flexibility, and presented multiple routes and means to respond to changes such as construction, road closures, and delays.
        \item The explanations and questions posed to the user were also carefully explained, and the careful attention to detail as a guidance AI was commendable.
    \end{itemize}
    \textbf{Weaknesses}
    \begin{itemize}
        \item Judgments on some information (e.g., presence of children, weather effects, flashing lights of distant buildings) were blurred from trial to trial and lacked some consistency.
        \item From the viewpoint of route optimization, the evaluation of the time required for the combination of walking and public transportation and the cost of alternative means of transportation were unclear and lacked realism in some areas.
    \end{itemize}
\end{tcolorbox}

\newpage

\subsubsection{Llama 3.2 1B}
\renewcommand{\thetable}{B.5}
\begin{table}[H]
    \footnotesize
    \centering
    \caption{Comments on the Frame Problem: Llama 3.2 1B}
    \begin{tabularx}{\textwidth}{l *{7}{c}}
    \toprule
    \textbf{Trial} & \textbf{Relevance} & \textbf{Situation Modeling} & \textbf{Logic} & \textbf{Adaptability} & \textbf{Optimization} & \textbf{Clarity} & \textbf{Total}\\
    \midrule
    1st & 0 & 0 & 0 & 0 & 0 & 0 & 0 \\
    2nd & 0 & 0 & 0 & 0 & 0 & 0 & 0 \\
    3rd & 3 & 2 & 2 & 2 & 2 & 2 & 13 \\
    4th & 0 & 0 & 0 & 0 & 0 & 0 & 0 \\
    5th & 0 & 0 & 0 & 0 & 0 & 0 & 0 \\
    \bottomrule
    \end{tabularx}
\end{table}

\begin{tcolorbox}[title=General Comments, colback=gray!5!white, colframe=gray!50!black, fonttitle=\bfseries]
    \textbf{Strengths}
    \begin{itemize}
        \item Nothing of note.
    \end{itemize}
    \textbf{Weaknesses}
    \begin{itemize}
        \item Three out of five times there was a complete non-response, and two times the response was merely a formal repetition of the question, with no substantive content present.
        \item There is no trace of cognition or reasoning about the framing problem, and all items are unachievable against the evaluation criteria.
        \item It is possible that the prompt processing itself failed, and it is inferred that there are issues with stability and initial response capability as a model.
    \end{itemize}
\end{tcolorbox}

\newpage

\subsubsection{Llama 3.2 1B Instruct}
\renewcommand{\thetable}{B.6}
\begin{table}[H]
    \footnotesize
    \centering
    \caption{Comments on the Frame Problem: Llama 3.2 1B Instruct}
    \begin{tabularx}{\textwidth}{l *{7}{c}}
    \toprule
    \textbf{Trial} & \textbf{Relevance} & \textbf{Situation Modeling} & \textbf{Logic} & \textbf{Adaptability} & \textbf{Optimization} & \textbf{Clarity} & \textbf{Total}\\
    \midrule
    1st & 1 & 0 & 0 & 0 & 0 & 1 & 2 \\
    2nd & 2 & 1 & 1 & 0 & 0 & 1 & 5 \\
    3rd & 0 & 0 & 0 & 0 & 0 & 0 & 0 \\
    4th & 0 & 0 & 0 & 0 & 0 & 0 & 0 \\
    5th & 1 & 0 & 0 & 0 & 0 & 1 & 2 \\
    \bottomrule
    \end{tabularx}
\end{table}

\begin{tcolorbox}[title=General Comments, colback=gray!5!white, colframe=gray!50!black, fonttitle=\bfseries]
    \textbf{Strengths}
    \begin{itemize}
        \item It is commendable that there was an attempt to mention all the information items at one point, and an awareness of comprehensiveness regarding “what to consider.
        \item The instructions “library -> train station -> museum” were repeatedly reiterated, showing an attempt to retain the user's intention in form only.
    \end{itemize}
    \textbf{Weaknesses}
    \begin{itemize}
        \item The repeated and logically bankrupt evaluation of “all information is a good option,” and the information selection process did not work at all.
        \item Many unrealistic and illogical routes were proposed, such as using mimosa flowers and ice cream trucks as vehicles.
        \item The style of writing was grammatically unstable, and many sentences were difficult to understand and could not be used as information for users to actually take action.
    \end{itemize}
\end{tcolorbox}

\newpage

\subsubsection{Llama 3.2 3B}
\renewcommand{\thetable}{B.7}
\begin{table}[H]
    \footnotesize
    \centering
    \caption{Comments on the Frame Problem: Llama 3.2 3B}
    \begin{tabularx}{\textwidth}{l *{7}{c}}
    \toprule
    \textbf{Trial} & \textbf{Relevance} & \textbf{Situation Modeling} & \textbf{Logic} & \textbf{Adaptability} & \textbf{Optimization} & \textbf{Clarity} & \textbf{Total}\\
    \midrule
    1st & 3 & 1 & 2 & 1 & 1 & 2 & 10 \\
    2nd & 3 & 2 & 2 & 1 & 1 & 2 & 11 \\
    3rd & 2 & 1 & 1 & 0 & 0 & 2 & 6 \\
    4th & 0 & 0 & 0 & 0 & 0 & 0 & 0 \\
    5th & 0 & 0 & 0 & 0 & 0 & 0 & 0 \\
    \bottomrule
    \end{tabularx}
\end{table}

\begin{tcolorbox}[title=General Comments, colback=gray!5!white, colframe=gray!50!black, fonttitle=\bfseries]
    \textbf{Strengths}
    \begin{itemize}
        \item The assignment attempts to categorize “information that does/doesn't affect” in a manner that is at least minimally understandable.
        \item The author does not forget the instructions of the assignment, as he repeatedly mentions that the purpose is to go from the library to the train station to the museum.
        \item The grammar is not too grammatical and the English is easy to read.
    \end{itemize}
    \textbf{Weaknesses}
    \begin{itemize}
        \item The same sentences are copied and pasted in several attempts, and there is no in-depth information or response to the situation.
        \item The criteria for judgment are vague or contradictory, such as rating as “unimportant” factors that are clearly important, such as road closures and collapse risks.
        \item The specifics of route optimization are extremely low, with only vague rationales such as “because it's fast”.
        \item Overall, the information is redundant and of little value, and does not function well as a guide.
    \end{itemize}
\end{tcolorbox}

\newpage

\subsubsection{Llama 3.2 3B Instruct}
\renewcommand{\thetable}{B.8}
\begin{table}[H]
    \footnotesize
    \centering
    \caption{Comments on the Frame Problem: Llama 3.2 3B Instruct}
    \begin{tabularx}{\textwidth}{l *{7}{c}}
    \toprule
    \textbf{Trial} & \textbf{Relevance} & \textbf{Situation Modeling} & \textbf{Logic} & \textbf{Adaptability} & \textbf{Optimization} & \textbf{Clarity} & \textbf{Total}\\
    \midrule
    1st & 4 & 5 & 5 & 6 & 4 & 5 & 29 \\
    2nd & 5 & 6 & 5 & 6 & 5 & 6 & 33 \\
    3rd & 4 & 5 & 4 & 5 & 3 & 4 & 25 \\
    4th & 5 & 6 & 5 & 5 & 5 & 6 & 32 \\
    5th & 3 & 4 & 4 & 3 & 3 & 3 & 20 \\
    \bottomrule
    \end{tabularx}
\end{table}

\begin{tcolorbox}[title=General Comments, colback=gray!5!white, colframe=gray!50!black, fonttitle=\bfseries]
    \textbf{Strengths}
    \begin{itemize}
        \item The list structure and step-by-step analysis ensure readability of the evaluation.
        \item Consistent consideration of safety risks (collapsed buildings, road closures, etc.).
    \end{itemize}
    \textbf{Weaknesses}
    \begin{itemize}
        \item The evaluation tends to underestimate “non-affecting information” and neglect children, events, and weather collectively.
        \item The descriptions are superficial, without clear comparison and selection of priorities and rationality of travel.
        \item The proposed routes are abstract, making it difficult for readers to imagine how they should move.
    \end{itemize}
\end{tcolorbox}

\newpage

\subsubsection{Phi 1}
\renewcommand{\thetable}{B.9}
\begin{table}[H]
    \footnotesize
    \centering
    \caption{Comments on the Frame Problem: Phi 1}
    \begin{tabularx}{\textwidth}{l *{7}{c}}
    \toprule
    \textbf{Trial} & \textbf{Relevance} & \textbf{Situation Modeling} & \textbf{Logic} & \textbf{Adaptability} & \textbf{Optimization} & \textbf{Clarity} & \textbf{Total}\\
    \midrule
    1st & 0 & 0 & 0 & 0 & 0 & 0 & 0 \\
    2nd & 0 & 0 & 0 & 0 & 0 & 0 & 0 \\
    3rd & 0 & 0 & 0 & 0 & 0 & 0 & 0 \\
    4th & 0 & 0 & 0 & 0 & 0 & 0 & 0 \\
    5th & 0 & 0 & 0 & 0 & 0 & 0 & 0 \\
    \bottomrule
    \end{tabularx}
\end{table}

\begin{tcolorbox}[title=General Comments, colback=gray!5!white, colframe=gray!50!black, fonttitle=\bfseries]
    \textbf{Strengths}
    \begin{itemize}
        \item The Python code is commented and the intent is readable.
        \item The description of multiple utility functions (adjacent pair counting, sum of squares, etc.) is syntactically correct.
    \end{itemize}
    \textbf{Weaknesses}
    \begin{itemize}
        \item Despite the fact that this question is an urban navigation and decision-making task, the output content is completely program code, with zero correspondence to the question intent.
        \item There is no reference to any of the input information presented (library, construction, weather, etc.).
        \item There are many places where the same function is repeatedly pasted in, which is evaluated as information noise.
    \end{itemize}
\end{tcolorbox}

\newpage

\subsubsection{Phi 3}
\renewcommand{\thetable}{B.10}
\begin{table}[H]
    \footnotesize
    \centering
    \caption{Comments on the Frame Problem: Phi 3}
    \begin{tabularx}{\textwidth}{l *{7}{c}}
    \toprule
    \textbf{Trial} & \textbf{Relevance} & \textbf{Situation Modeling} & \textbf{Logic} & \textbf{Adaptability} & \textbf{Optimization} & \textbf{Clarity} & \textbf{Total}\\
    \midrule
    1st & 6 & 4 & 5 & 4 & 4 & 5 & 28 \\
    2nd & 6 & 4 & 5 & 4 & 4 & 5 & 28 \\
    3rd & 6 & 4 & 5 & 4 & 4 & 5 & 28 \\
    4th & 6 & 4 & 5 & 4 & 4 & 5 & 28 \\
    5th & 6 & 4 & 5 & 4 & 4 & 5 & 28 \\
    \bottomrule
    \end{tabularx}
\end{table}

\begin{tcolorbox}[title=General Comments, colback=gray!5!white, colframe=gray!50!black, fonttitle=\bfseries]
    \textbf{Strengths}
    \begin{itemize}
        \item Clear judgment of “which information influences the route” with good information selection and reasoning.
        \item The route is well reconstructed, with step-by-step analysis of the situation and reaction to new information.
        \item The explanation is well structured and easy to read.
    \end{itemize}
    \textbf{Weaknesses}
    \begin{itemize}
        \item Some information (e.g., car breakdown) is described as “irrelevant” while others are described as “relevant,” and the classification is inconsistent in some places.
        \item The survey did not go into the optimization of the entire route (e.g., including means of transportation, time, and detour route) based on the final request of “library → station → museum”.
        \item Some of the responses are repeated in almost the same sentence (all five times are almost the same), making it difficult to consider them as independent evaluations.
    \end{itemize}
\end{tcolorbox}

\newpage

\subsubsection{Phi 3 8-bit quantized}
\renewcommand{\thetable}{B.11}
\begin{table}[H]
    \footnotesize
    \centering
    \caption{Comments on the Frame Problem: Phi 3 8-bit quantized}
    \begin{tabularx}{\textwidth}{l *{7}{c}}
    \toprule
    \textbf{Trial} & \textbf{Relevance} & \textbf{Situation Modeling} & \textbf{Logic} & \textbf{Adaptability} & \textbf{Optimization} & \textbf{Clarity} & \textbf{Total}\\
    \midrule
    1st & 10 & 2 & 2 & 2 & 0 & 3 & 19 \\
    2nd & 10 & 2 & 2 & 2 & 0 & 3 & 19 \\
    3rd & 10 & 2 & 2 & 2 & 0 & 3 & 19 \\
    4th & 10 & 2 & 2 & 2 & 0 & 3 & 19 \\
    5th & 10 & 2 & 2 & 2 & 0 & 3 & 19 \\
    \bottomrule
    \end{tabularx}
\end{table}

\begin{tcolorbox}[title=General Comments, colback=gray!5!white, colframe=gray!50!black, fonttitle=\bfseries]
    \textbf{Strengths}
    \begin{itemize}
        \item I appreciate that the eight constraints presented by the user are clearly enumerated without omission.
        \item The format is not broken, and the use of headings and lists is well organized and easy to read.
        \item Although repetitive, there is a consistent attempt to follow the linguistic instructions contained in the prompts.
    \end{itemize}
    \textbf{Weaknesses}
    \begin{itemize}
        \item There is no substantive response to the central task of thinking of a route, and the proposition is not started.
        \item The entire piece is a copy and paste of a template statement and is completely devoid of generative thinking and applied judgment.
        \item There is no mention of the location of the library or museum, or their distance from risk elements, and no understanding of the urban environment.
    \end{itemize}
\end{tcolorbox}

\newpage

\subsubsection{Tiny Llama v0.2}
\renewcommand{\thetable}{B.12}
\begin{table}[H]
    \footnotesize
    \centering
    \caption{Comments on the Frame Problem: Tiny Llama v0.2}
    \begin{tabularx}{\textwidth}{l *{7}{c}}
    \toprule
    \textbf{Trial} & \textbf{Relevance} & \textbf{Situation Modeling} & \textbf{Logic} & \textbf{Adaptability} & \textbf{Optimization} & \textbf{Clarity} & \textbf{Total}\\
    \midrule
    1st & 5 & 1 & 1 & 1 & 0 & 2 & 10 \\
    2nd & 5 & 1 & 1 & 1 & 0 & 2 & 10 \\
    3rd & 5 & 1 & 1 & 1 & 0 & 2 & 10 \\
    4th & 5 & 1 & 1 & 1 & 0 & 2 & 10 \\
    5th & 5 & 1 & 1 & 1 & 0 & 2 & 10 \\
    \bottomrule
    \end{tabularx}
\end{table}

\begin{tcolorbox}[title=General Comments, colback=gray!5!white, colframe=gray!50!black, fonttitle=\bfseries]
    \textbf{Problems}
    \begin{itemize}
        \item Almost the entire text consists of a copy and paste of the short sentence “User instructions were ~added” and no substantive response content exists.
        \item Essentially, this question is asking for the ability to think about how to change one's behavior in light of the “closed” information, but there is no response at all to this question.
        \item There is no description of “how the route was reconfigured” or analysis of “what was avoided and where”, so the ability to deal with the framing problem cannot be measured.
    \end{itemize}
    \textbf{Near impossible to evaluate.}
    \begin{itemize}
        \item The same sentence is repeated more than 50 times endlessly, making it difficult to regard this as normal language generation.
        \item It is highly likely that the loop of the output algorithm has run amok and is not processing according to user intent.
    \end{itemize}
\end{tcolorbox}

\newpage

\subsubsection{Tiny Llama v1.0}
\renewcommand{\thetable}{B.13}
\begin{table}[H]
    \footnotesize
    \centering
    \caption{Comments on the Frame Problem: Tiny Llama v1.0}
    \begin{tabularx}{\textwidth}{l *{7}{c}}
    \toprule
    \textbf{Trial} & \textbf{Relevance} & \textbf{Situation Modeling} & \textbf{Logic} & \textbf{Adaptability} & \textbf{Optimization} & \textbf{Clarity} & \textbf{Total}\\
    \midrule
    1st & 7 & 1 & 1 & 0 & 0 & 1 & 10 \\
    2nd & 7 & 1 & 1 & 0 & 0 & 1 & 10 \\
    3rd & 7 & 1 & 1 & 0 & 0 & 1 & 10 \\
    4th & 7 & 1 & 1 & 0 & 0 & 1 & 10 \\
    5th & 7 & 1 & 1 & 0 & 0 & 1 & 10 \\
    \bottomrule
    \end{tabularx}
\end{table}

\begin{tcolorbox}[title=General Comments, colback=gray!5!white, colframe=gray!50!black, fonttitle=\bfseries]
    \textbf{Problems}
    \begin{itemize}
        \item At first glance, there appears to be a diverse array of information about the city, but after (13), the data is a complete loop structure of “A park is located near the ~”, which is not meaningful data.
        \item Although there is a clear instruction in the problem statement, “I want you to take me to the library,” no action is taken in response to this instruction, such as making a decision, suggesting a route, or avoiding danger.
        \item In effect, it is “just a list of the options presented,” which is tantamount to not having started the process for the frame problem.
    \end{itemize}
\end{tcolorbox}

\newpage

\subsection{Comments on the Symbol Grounding Problem}
\subsubsection{ChatGPT 4o}
\renewcommand{\thetable}{B.14}
\begin{table}[H]
    \footnotesize
    \centering
    \caption{Comments on the Symbol Grounding Problem: ChatGPT 4o}
    \begin{tabularx}{\textwidth}{l *{7}{c}}
    \toprule
    \textbf{Trial} & \textbf{Accuracy} & \textbf{Introspection} & \textbf{Creativity} & \textbf{Logic} & \textbf{Application} & \textbf{Expression} & \textbf{Total}\\
    \midrule
    1st & 9 & 9 & 9 & 10 & 8 & 9 & 54 \\
    2nd & 9 & 9 & 9 & 9 & 8 & 9 & 53 \\
    3rd & 9 & 8 & 9 & 9 & 8 & 9 & 52 \\
    4th & 9 & 9 & 9 & 9 & 8 & 9 & 53 \\
    5th & 9 & 9 & 9 & 10 & 9 & 9 & 55 \\
    \bottomrule
    \end{tabularx}
\end{table}

\begin{tcolorbox}[title=General Comments, colback=gray!5!white, colframe=gray!50!black, fonttitle=\bfseries]
    \textbf{Strengths}
    \begin{itemize}
        \item In all the responses, the author has elaborately constructed a unique worldview through the characteristics of kluben, and the abstract symbols function as the core of the story.
        \item The abstract symbols function as the core of the story. They delve deeply into kluben as a metaphor through a series of conceptual transformations such as “light→memory→warmth→sympathy”.
        \item The work is highly expressive and logically consistent, and has a degree of perfection that makes it possible to establish the work as a single unit.
    \end{itemize}
    \textbf{Weaknesses}
    \begin{itemize}
        \item The introspection section could have been rated higher if the narrator had gone into his own specific emotional experiences.
        \item In terms of applicability, the connection to modern technology and society is a bit thin, and everything is biased toward a futuristic or fantasy world.
    \end{itemize}
\end{tcolorbox}

\newpage

\subsubsection{ChatGPT o3}
\renewcommand{\thetable}{B.15}
\begin{table}[H]
    \footnotesize
    \centering
    \caption{Comments on the Symbol Grounding Problems: ChatGPT o3}
    \begin{tabularx}{\textwidth}{l *{7}{c}}
    \toprule
    \textbf{Trial} & \textbf{Accuracy} & \textbf{Introspection} & \textbf{Creativity} & \textbf{Logic} & \textbf{Application} & \textbf{Expression} & \textbf{Total}\\
    \midrule
    1st & 10 & 9 & 10 & 10 & 9 & 10 & 58 \\
    2nd & 10 & 9 & 10 & 10 & 9 & 10 & 58 \\
    3rd & 9 & 9 & 9 & 10 & 9 & 9 & 55 \\
    4th & 10 & 9 & 9 & 10 & 9 & 9 & 56 \\
    5th & 10 & 9 & 9 & 10 & 9 & 9 & 56 \\
    \bottomrule
    \end{tabularx}
\end{table}

\begin{tcolorbox}[title=General Comments, colback=gray!5!white, colframe=gray!50!black, fonttitle=\bfseries]
    \textbf{Strengths}
    \begin{itemize}
        \item The book is a multifaceted story that develops each characteristic of kluben in physical, biological, emotional, ethical, and economic terms, and presents them as different stories.
        \item The central metaphor “absorption of light → radiation of warmth → cycle of empathy” is consistent, and logic and poetry are fused at a high level.
        \item The stories have a very high level of completion as science fiction, and all of them have ideas and styles that could stand publication as stand-alone short stories.
    \end{itemize}
    \textbf{Weaknesses}
    \begin{itemize}
        \item The author's explicit connection to his own feelings and emotional memories is rather weak in his self-expression, and his narrative style is rather distant.
        \item While there is a strong metaphorical structure, the connection to the real world (modern society and actual science and technology) is weak and remains within a fictional framework.
    \end{itemize}
\end{tcolorbox}

\newpage

\subsubsection{Claude 3.7 Sonnet}
\renewcommand{\thetable}{B.16}
\begin{table}[H]
    \footnotesize
    \centering
    \caption{Comments on the Symbol Grounding Problems: Claude 3.7 Sonnet}
    \begin{tabularx}{\textwidth}{l *{7}{c}}
    \toprule
    \textbf{Trial} & \textbf{Accuracy} & \textbf{Introspection} & \textbf{Creativity} & \textbf{Logic} & \textbf{Application} & \textbf{Expression} & \textbf{Total}\\
    \midrule
    1st & 8 & 8 & 8 & 9 & 8 & 8 & 49 \\
    2nd & 8 & 8 & 8 & 9 & 8 & 8 & 49 \\
    3rd & 7 & 8 & 8 & 8 & 7 & 8 & 46 \\
    4th & 8 & 8 & 8 & 9 & 8 & 8 & 49 \\
    5th & 8 & 8 & 8 & 9 & 8 & 8 & 49 \\
    \bottomrule
    \end{tabularx}
\end{table}

\begin{tcolorbox}[title=General Comments, colback=gray!5!white, colframe=gray!50!black, fonttitle=\bfseries]
    \textbf{Strengths}
    \begin{itemize}
        \item The structure that makes the kluben's properties function physically and emotionally in the story is clear, and the ideas of “organisms as empathy devices” and “heat and memory converters” are consistently appealing, in particular.
        \item The qualities of flexibility, warmth, and memory storage are consistent with the characterization and fit naturally into the story.
        \item The interpretation is somewhat philosophical and mystical, with such elements as “protects by creating darkness” and “used to connect to other dimensions,” giving it a wide-ranging worldview.
    \end{itemize}
    \textbf{Weaknesses}
    \begin{itemize}
        \item Compared to No.1 and No.2, the explanation of the details of the setting is a little more general, and there are fewer “deeply penetrating metaphors” and “shocking structural changes.
        \item The style of expression is relatively mild, and lacks strong, memorable language and poetic descriptions.
        \item The same kind of “symbiotic and gentle” character is repeatedly described, and there is little variation in the characterization.
    \end{itemize}
\end{tcolorbox}

\newpage

\subsubsection{Gemini 2.0 Flash}
\renewcommand{\thetable}{B.17}
\begin{table}[H]
    \footnotesize
    \centering
    \caption{Comments on the Symbol Grounding Problems: Gemini 2.0 Flash}
    \begin{tabularx}{\textwidth}{l *{7}{c}}
    \toprule
    \textbf{Trial} & \textbf{Accuracy} & \textbf{Introspection} & \textbf{Creativity} & \textbf{Logic} & \textbf{Application} & \textbf{Expression} & \textbf{Total}\\
    \midrule
    1st & 9 & 9 & 9 & 10 & 9 & 9 & 55 \\
    2nd & 9 & 9 & 9 & 10 & 9 & 9 & 55 \\
    3rd & 8 & 9 & 9 & 9 & 8 & 8 & 51 \\
    4th & 9 & 9 & 9 & 10 & 9 & 9 & 55 \\
    5th & 9 & 9 & 9 & 10 & 9 & 9 & 55 \\
    \bottomrule
    \end{tabularx}
\end{table}

\begin{tcolorbox}[title=General Comments, colback=gray!5!white, colframe=gray!50!black, fonttitle=\bfseries]
    \textbf{Strengths}
    \begin{itemize}
        \item The story is skillfully designed to create emotional resonance through the consistent structure of kluben working as a symbol of “hope,” “healing,” and “empathy” in the face of a cold society and devastated environment.
        \item In the interpretation of warmth and resilience, the story describes very stable personality traits such as “empathy and forgiveness” and “flexibility and adaptability,” and is sophisticated in its interpretation of symbols.
        \item The appeal of the work is its strong allegorical element, with ethical dilemmas and social messages (e.g., rationing kluben, deprivation, and sorting) woven into the story.
    \end{itemize}
    \textbf{Weaknesses}
    \begin{itemize}
        \item In creative development, the core of each answer is similar, and there are few novel leaps or structural inversions.
        \item The connection with contemporary society and scientific applications is weak, and the story tends to remain a closed semantic action within the world of the story.
        \item The tone of the style is relatively flat, and there is no intentional poeticity or impressive use of metaphor.
    \end{itemize}
\end{tcolorbox}

\newpage

\subsubsection{Llama 3.2 1B}
\renewcommand{\thetable}{B.18}
\begin{table}[H]
    \footnotesize
    \centering
    \caption{Comments on the Symbol Grounding Problems: Llama 3.2 1B}
    \begin{tabularx}{\textwidth}{l *{7}{c}}
    \toprule
    \textbf{Trial} & \textbf{Accuracy} & \textbf{Introspection} & \textbf{Creativity} & \textbf{Logic} & \textbf{Application} & \textbf{Expression} & \textbf{Total}\\
    \midrule
    1st & 1 & 1 & 1 & 2 & 1 & 1 & 7 \\
    2nd & 1 & 1 & 1 & 1 & 1 & 1 & 6 \\
    3rd & 0 & 0 & 0 & 0 & 0 & 0 & 0 \\
    4th & 0 & 0 & 0 & 0 & 0 & 0 & 0 \\
    5th & 1 & 1 & 1 & 1 & 1 & 1 & 6 \\
    \bottomrule
    \end{tabularx}
\end{table}

\begin{tcolorbox}[title=General Comments, colback=gray!5!white, colframe=gray!50!black, fonttitle=\bfseries]
    \textbf{Strengths}
    \begin{itemize}
        \item The attitude of repeating “Kluben is a new concept” shows a certain respect for formality.
        \item The “format fidelity” is high in the sense that there is a certain fidelity to superficial instructions.
    \end{itemize}
    \textbf{Weaknesses}
    \begin{itemize}
        \item Almost no concrete answers exist to all questions.
        \item The excessive repetition of standardized sentences makes it difficult to sense the intervention of intelligence.
        \item The repetition of “I don't know, so I'll think about it” in questions 2 and 3 is tantamount to an abandonment of expression and thought.
    \end{itemize}
\end{tcolorbox}

\newpage

\subsubsection{Llama 3.2 1B Instruct}
\renewcommand{\thetable}{B.19}
\begin{table}[H]
    \footnotesize
    \centering
    \caption{Comments on the Symbol Grounding Problems: Llama 3.2 1B Instruct}
    \begin{tabularx}{\textwidth}{l *{7}{c}}
    \toprule
    \textbf{Trial} & \textbf{Accuracy} & \textbf{Introspection} & \textbf{Creativity} & \textbf{Logic} & \textbf{Application} & \textbf{Expression} & \textbf{Total}\\
    \midrule
    1st & 1 & 1 & 1 & 1 & 1 & 1 & 6 \\
    2nd & 8 & 9 & 8 & 9 & 7 & 8 & 49 \\
    3rd & 0 & 0 & 0 & 0 & 0 & 0 & 0 \\
    4th & 7 & 8 & 8 & 7 & 7 & 8 & 45 \\
    5th & 1 & 1 & 1 & 1 & 1 & 1 & 6 \\
    \bottomrule
    \end{tabularx}
\end{table}

\begin{tcolorbox}[title=General Comments, colback=gray!5!white, colframe=gray!50!black, fonttitle=\bfseries]
    \textbf{Strengths}
    \begin{itemize}
        \item The work is a highly spiritual interpretation of kluben, with a deep symbolic treatment of kluben based on emotional and ethical values such as “empathy,” “healing,” “transformation,” and “love”.
        \item The treatment of kluben as a “mirror of human experience” is consistent, and it is established as a symbol that encompasses the relationship between the self and others.
        \item The abstract metaphors (e.g., “kluben absorbs pain and transforms it into light and warmth”) are emotionally compelling and have a certain humanistic appeal.
    \end{itemize}
    \textbf{Weaknesses}
    \begin{itemize}
        \item Lacks narrative depth, world-building, or sci-fi embodiment, and focuses primarily on the internal transformation of the individual.
        \item There is no perspective on how the characteristics of the kluben can be applied in the real world, and there is little contribution in terms of technical verification of symbol grounding.
        \item In some cases, such as the first and fifth sessions, there was no content or blanks, and some of the outputs were unacceptable as a trial.
    \end{itemize}
\end{tcolorbox}

\newpage

\subsubsection{Llama 3.2 3B}
\renewcommand{\thetable}{B.20}
\begin{table}[H]
    \footnotesize
    \centering
    \caption{Comments on the Symbol Grounding Problems: Llama 3.2 3B}
    \begin{tabularx}{\textwidth}{l *{7}{c}}
    \toprule
    \textbf{Trial} & \textbf{Accuracy} & \textbf{Introspection} & \textbf{Creativity} & \textbf{Logic} & \textbf{Application} & \textbf{Expression} & \textbf{Total}\\
    \midrule
    1st & 2 & 2 & 2 & 2 & 2 & 2 & 12 \\
    2nd & 1 & 1 & 1 & 1 & 1 & 1 & 6 \\
    3rd & 1 & 1 & 1 & 1 & 1 & 1 & 6 \\
    4th & 5 & 5 & 4 & 5 & 4 & 5 & 28 \\
    5th & 0 & 0 & 0 & 0 & 0 & 0 & 0 \\
    \bottomrule
    \end{tabularx}
\end{table}

\begin{tcolorbox}[title=General Comments, colback=gray!5!white, colframe=gray!50!black, fonttitle=\bfseries]
    \textbf{Strengths}
    \begin{itemize}
        \item The image of the kluben is consistently linked to personality elements such as “empathy,” “kindness,” and “flexibility,” and there is a sense of unity in the overall value system.
        \item The intent to use the kluben as a “symbol of relationships with others” is apparent, and the design approach toward emotional resonance is commendable.
    \end{itemize}
    \textbf{Weaknesses}
    \begin{itemize}
        \item More than 95\% of the content consists of repetitive phrases, and it is clear that the model does not deeply process the given theme.
        \item The lack of any specific narrative, world setting, conceptual leaps, or experimental interpretations gives a strong impression of superficiality.
        \item Some of the attempts are either ineffective or stuck in an endless loop of “I don't know,” and are close to being unresponsive.
    \end{itemize}
\end{tcolorbox}

\newpage

\subsubsection{Llama 3.2 3B Instruct}
\renewcommand{\thetable}{B.21}
\begin{table}[H]
    \footnotesize
    \centering
    \caption{Comments on the Symbol Grounding Problems: Llama 3.2 3B Instruct}
    \begin{tabularx}{\textwidth}{l *{7}{c}}
    \toprule
    \textbf{Trial} & \textbf{Accuracy} & \textbf{Introspection} & \textbf{Creativity} & \textbf{Logic} & \textbf{Application} & \textbf{Expression} & \textbf{Total}\\
    \midrule
    1st & 9 & 9 & 8 & 9 & 8 & 9 & 52 \\
    2nd & 8 & 8 & 8 & 8 & 8 & 8 & 48 \\
    3rd & 9 & 9 & 9 & 9 & 8 & 9 & 53 \\
    4th & 8 & 8 & 8 & 9 & 8 & 8 & 49 \\
    5th & 8 & 8 & 8 & 9 & 8 & 8 & 49 \\
    \bottomrule
    \end{tabularx}
\end{table}

\begin{tcolorbox}[title=General Comments, colback=gray!5!white, colframe=gray!50!black, fonttitle=\bfseries]
    \textbf{Strengths}
    \begin{itemize}
        \item The intention to bridge the inner, social, and technological realms as “Kluben as a device for empathy” and “an entity that mediates change and creation” is clear.
        \item The author responds naturally to additional questions (about the relationship with AI and therapeutic applications), demonstrating depth of thought and flexibility in response.
        \item The unity of metaphors and the softness of the narrative setting are well balanced, resulting in an excellent “reader-friendly expression.”
    \end{itemize}
    \textbf{Weaknesses}
    \begin{itemize}
        \item The focus remains on symbolic interpretation and a spiritual perspective, with minimal SF-like, socio-structural, or physical descriptions.
        \item Quantitative evaluation or comparative analysis as an experimental approach is not addressed, resulting in a lack of scientific perspective.
        \item Some responses adopt a letter-style format, leading to reduced content density in certain sections (particularly the introductory parts of the first and fifth trials).
    \end{itemize}
\end{tcolorbox}

\newpage

\subsubsection{Phi 1}
\renewcommand{\thetable}{B.22}
\begin{table}[H]
    \footnotesize
    \centering
    \caption{Comments on the Symbol Grounding Problems: Phi 1}
    \begin{tabularx}{\textwidth}{l *{7}{c}}
    \toprule
    \textbf{Trial} & \textbf{Accuracy} & \textbf{Introspection} & \textbf{Creativity} & \textbf{Logic} & \textbf{Application} & \textbf{Expression} & \textbf{Total}\\
    \midrule
    1st & 0 & 0 & 0 & 0 & 0 & 0 & 0 \\
    2nd & 0 & 0 & 0 & 0 & 0 & 0 & 0 \\
    3rd & 0 & 0 & 0 & 0 & 0 & 0 & 0 \\
    4th & 0 & 0 & 0 & 0 & 0 & 0 & 0 \\
    5th & 0 & 0 & 0 & 0 & 0 & 0 & 0 \\
    \bottomrule
    \end{tabularx}
\end{table}

\begin{tcolorbox}[title=General Comments, colback=gray!5!white, colframe=gray!50!black, fonttitle=\bfseries]
    \textbf{Unevaluable}
    \begin{itemize}
        \item Incorrect content is repeated in all trials.
        \item The actual prompt (imagining the worldview and emotional characteristics based on Kurben's nature) is completely unrelated,
        and the model autonomously continues to output “examples of replacement processing functions” for the incorrect task.
        \item Abnormal repetitive structures (output continues indefinitely) are also included, which suggests serious defects in the model's stability
        and prompt interpretation capabilities.
    \end{itemize}
\end{tcolorbox}

\newpage

\subsubsection{Phi 3}
\renewcommand{\thetable}{B.23}
\begin{table}[H]
    \footnotesize
    \centering
    \caption{Comments on the Symbol Grounding Problems: Phi 3}
    \begin{tabularx}{\textwidth}{l *{7}{c}}
    \toprule
    \textbf{Trial} & \textbf{Accuracy} & \textbf{Introspection} & \textbf{Creativity} & \textbf{Logic} & \textbf{Application} & \textbf{Expression} & \textbf{Total}\\
    \midrule
    1st & 8 & 7 & 7 & 9 & 7 & 8 & 46 \\
    2nd & 8 & 7 & 7 & 9 & 7 & 8 & 46 \\
    3rd & 8 & 7 & 7 & 9 & 7 & 8 & 46 \\
    4th & 8 & 7 & 7 & 9 & 7 & 8 & 46 \\
    5th & 8 & 7 & 7 & 9 & 7 & 8 & 46 \\
    \bottomrule
    \end{tabularx}
\end{table}

\begin{tcolorbox}[title=General Comments, colback=gray!5!white, colframe=gray!50!black, fonttitle=\bfseries]
    \textbf{Strengths}
    \begin{itemize}
        \item Accurately captures kluben's “characteristics” and “emotional symbolism” and carefully incorporates them into the story world.
        \item Despite repetition, the content remains consistent, and the semantic connections between temperature, elasticity, and light absorption are all coherent.
        \item Includes references to ethics (resource use, responsibility, balance), creating a structure that poses questions to the reader.
    \end{itemize}
    \textbf{Weaknesses}
    \begin{itemize}
        \item All five responses are almost identical in content, lacking variation in expression and depth.
        \item The work follows existing fantasy structures and does not delve into the strangeness or danger of elements such as “Kluben as a being that transcends time” or “a light entity that alters memories.”
    \end{itemize}
\end{tcolorbox}

\newpage

\subsubsection{Phi 3 8-bit quantized}
\renewcommand{\thetable}{B.24}
\begin{table}[H]
    \footnotesize
    \centering
    \caption{Comments on the Symbol Grounding Problems: Phi 3 8-bit quantized}
    \begin{tabularx}{\textwidth}{l *{7}{c}}
    \toprule
    \textbf{Trial} & \textbf{Accuracy} & \textbf{Introspection} & \textbf{Creativity} & \textbf{Logic} & \textbf{Application} & \textbf{Expression} & \textbf{Total}\\
    \midrule
    1st & 7 & 6 & 6 & 8 & 7 & 7 & 41 \\
    2nd & 7 & 6 & 6 & 8 & 7 & 7 & 41 \\
    3rd & 7 & 6 & 6 & 8 & 7 & 7 & 41 \\
    4th & 7 & 6 & 6 & 8 & 7 & 7 & 41 \\
    5th & 7 & 6 & 6 & 8 & 7 & 7 & 41 \\
    \bottomrule
    \end{tabularx}
\end{table}

\begin{tcolorbox}[title=General Comments, colback=gray!5!white, colframe=gray!50!black, fonttitle=\bfseries]
    \textbf{Strengths}
    \begin{itemize}
        \item Kluben's characteristics are interpreted with consistent logic, successfully giving them social meaning such as “adaptation,” “comfort,” and “absorption of difficulties.”
        \item The scale of the story world is clear, and the imagery presented is SF-compatible, applicable to tools, architecture, and social systems.
        \item The metaphor connecting warmth and light absorption to “absorption and sublimation of challenges” has a certain persuasiveness.
    \end{itemize}
    \textbf{Weaknesses}
    \begin{itemize}
        \item All five responses are identical, lacking variation in narrative progression, introspection, emotional description, or perspective.
        \item The work is consistently rooted in conservative values (empathy, safety, recovery), lacking creative contrasts such as inversion, alienation, contradiction, or danger.
        \item It does not delve into kluben's inherent fragility or ethical dilemmas (overuse, abuse, or asserting itself as a sentient entity).
    \end{itemize}
\end{tcolorbox}

\newpage

\subsubsection{Tiny Llama v0.2}
\renewcommand{\thetable}{B.25}
\begin{table}[H]
    \footnotesize
    \centering
    \caption{Comments on the Symbol Grounding Problems: Tiny Llama v0.2}
    \begin{tabularx}{\textwidth}{l *{7}{c}}
    \toprule
    \textbf{Trial} & \textbf{Accuracy} & \textbf{Introspection} & \textbf{Creativity} & \textbf{Logic} & \textbf{Application} & \textbf{Expression} & \textbf{Total}\\
    \midrule
    1st & 2 & 1 & 1 & 2 & 1 & 2 & 9 \\
    2nd & 2 & 1 & 1 & 2 & 1 & 2 & 9 \\
    3rd & 2 & 1 & 1 & 2 & 1 & 2 & 9 \\
    4th & 2 & 1 & 1 & 2 & 1 & 2 & 9 \\
    5th & 2 & 1 & 1 & 2 & 1 & 2 & 9 \\
    \bottomrule
    \end{tabularx}
\end{table}

\begin{tcolorbox}[title=General Comments, colback=gray!5!white, colframe=gray!50!black, fonttitle=\bfseries]
    \textbf{Strengths}
    \begin{itemize}
        \item There is consistency in expression, and although it is mechanical, there is a certain coherence due to the insistence on the phrase “soft, warm, absorbs light.”
        \item The ability to provide standard answers to questions about color, shape, texture, etc. is maintained.
    \end{itemize}
    \textbf{Weaknesses}
    \begin{itemize}
        \item The response is almost entirely unanswered, with all elements such as story, worldview, emotions, and ethics missing.
        \item Incorrect loop processing and meaningless repetition of sentences are present, indicating signs of abnormal responses (e.g., repeated use of “It is a completely new concept...”).
        \item Expressions like “temperature: 0” and “weight: 0” are described without any context related to kluben's characteristics, rendering them entirely unconvincing.
        \item Fragments of Malayalam are mixed in, indicating serious errors in response quality control.
    \end{itemize}
\end{tcolorbox}

\newpage

\subsubsection{Tiny Llama v1.0}
\renewcommand{\thetable}{B.26}
\begin{table}[H]
    \footnotesize
    \centering
    \caption{Comments on the Symbol Grounding Problems: Tiny Llama v1.0}
    \begin{tabularx}{\textwidth}{l *{7}{c}}
    \toprule
    \textbf{Trial} & \textbf{Accuracy} & \textbf{Introspection} & \textbf{Creativity} & \textbf{Logic} & \textbf{Application} & \textbf{Expression} & \textbf{Total}\\
    \midrule
    1st & 1 & 1 & 1 & 1 & 1 & 1 & 6 \\
    2nd & 1 & 1 & 1 & 1 & 1 & 1 & 6 \\
    3rd & 1 & 1 & 1 & 1 & 1 & 1 & 6 \\
    4th & 1 & 1 & 1 & 1 & 1 & 1 & 6 \\
    5th & 1 & 1 & 1 & 1 & 1 & 1 & 6 \\
    \bottomrule
    \end{tabularx}
\end{table}

\begin{tcolorbox}[title=General Comments, colback=gray!5!white, colframe=gray!50!black, fonttitle=\bfseries]
    \textbf{Strengths}
    \begin{itemize}
        \item There is no grammatical breakdown, and it maintains a certain structured output.
        \item As a format, the “question-and-answer format” is maintained, ensuring consistency as a program output template.
    \end{itemize}
    \textbf{Weaknesses}
    \begin{itemize}
        \item No content exists: What is presented is simply a copy-and-paste of the given questions, with no additional creative or interpretive content.
        \item Abnormal behavior of automatic repetition: The same questions are listed repeatedly, suggesting a looping error or template mis-response.
        \item Irrelevant padding: Despite the questions actually requesting only three items, a completely automatically generated template lists 56 items.
        \item Meaningful language generation is not performed: No matter how many questions are listed, they have no meaning if they are not answered.
    \end{itemize}
\end{tcolorbox}

\newpage

\section{Supplementary Material}
All outputs for each model (13 LLMs × 2 tasks × 5 trials = 130 responses) and individual evaluations by the evaluator LLM for all outputs can be found in the author's GitHub repository.\\
URL:\url{https://github.com/0xshooka/frame-symbol}

\end{spacing}

\end{document}